\documentclass[review,12pt,authoryear]{elsarticle}

\usepackage{amssymb}
\usepackage{amsmath}

\RequirePackage{iftex}
\ifPDFTeX\else
  \newcount\pdfcompresslevel
  \newcount\pdfoptionpdfminorversion
  \newcount\pdfgentounicode
  \providecommand{\pdfcatalog}[1]{}
  \providecommand{\pdfglyphtounicode}[2]{}
\fi

\usepackage{xcolor}
\definecolor{cmlabPrefix}{HTML}{0D207F}  
\definecolor{cmlabNumber}{HTML}{0D207F}  
\definecolor{cmlabCite}{HTML}{0D207F}    
\definecolor{cmlabURL}{HTML}{0D207F}     

\usepackage{hyperref}
\hypersetup{
  colorlinks=true,
  linkcolor=cmlabNumber,    
  citecolor=cmlabCite,      
  urlcolor=cmlabURL,        
  filecolor=cmlabURL,
  pdfborder={0 0 0},
}

\usepackage[capitalize,noabbrev]{cleveref}

\newcommand{\cmlabRef}[4]{%
  {\color{cmlabPrefix}#1}~#3{\color{cmlabNumber}#2}#4%
}
\newcommand{\cmlabRefParen}[4]{
  {\color{cmlabPrefix}#1}~#3{\color{cmlabNumber}(#2)}#4%
}

\crefformat{figure}     {\cmlabRef{Fig.}{#1}{#2}{#3}}
\Crefformat{figure}     {\cmlabRef{Figure}{#1}{#2}{#3}}
\crefrangeformat{figure}%
  {{\color{cmlabPrefix}Figs.}~#3{\color{cmlabNumber}#1}#4--#5{\color{cmlabNumber}#2}#6}
\crefmultiformat{figure}%
  {\cmlabRef{Figs.}{#1}{#2}{#3}}%
  {, #2{\color{cmlabNumber}#1}#3}%
  {, #2{\color{cmlabNumber}#1}#3}%
  {, and #2{\color{cmlabNumber}#1}#3}

\crefformat{table}      {\cmlabRef{Tab.}{#1}{#2}{#3}}
\Crefformat{table}      {\cmlabRef{Table}{#1}{#2}{#3}}
\crefrangeformat{table}%
  {{\color{cmlabPrefix}Tabs.}~#3{\color{cmlabNumber}#1}#4--#5{\color{cmlabNumber}#2}#6}
\crefmultiformat{table}%
  {\cmlabRef{Tabs.}{#1}{#2}{#3}}%
  {, #2{\color{cmlabNumber}#1}#3}%
  {, #2{\color{cmlabNumber}#1}#3}%
  {, and #2{\color{cmlabNumber}#1}#3}

\crefformat{section}    {\cmlabRef{Sec.}{#1}{#2}{#3}}
\Crefformat{section}    {\cmlabRef{Section}{#1}{#2}{#3}}
\crefformat{subsection} {\cmlabRef{Sec.}{#1}{#2}{#3}}
\Crefformat{subsection} {\cmlabRef{Section}{#1}{#2}{#3}}
\crefformat{subsubsection}{\cmlabRef{Sec.}{#1}{#2}{#3}}
\Crefformat{subsubsection}{\cmlabRef{Section}{#1}{#2}{#3}}
\crefrangeformat{section}%
  {{\color{cmlabPrefix}Secs.}~#3{\color{cmlabNumber}#1}#4--#5{\color{cmlabNumber}#2}#6}
\crefmultiformat{section}%
  {\cmlabRef{Secs.}{#1}{#2}{#3}}%
  {, #2{\color{cmlabNumber}#1}#3}%
  {, #2{\color{cmlabNumber}#1}#3}%
  {, and #2{\color{cmlabNumber}#1}#3}

\crefformat{equation}   {\cmlabRefParen{Eq.}{#1}{#2}{#3}}
\Crefformat{equation}   {\cmlabRefParen{Equation}{#1}{#2}{#3}}
\crefrangeformat{equation}%
  {{\color{cmlabPrefix}Eqs.}~#3{\color{cmlabNumber}(#1)}#4--#5{\color{cmlabNumber}(#2)}#6}
\crefmultiformat{equation}%
  {\cmlabRefParen{Eqs.}{#1}{#2}{#3}}%
  {, #2{\color{cmlabNumber}(#1)}#3}%
  {, #2{\color{cmlabNumber}(#1)}#3}%
  {, and #2{\color{cmlabNumber}(#1)}#3}

\crefname{algorithm}{Algo.}{Algos.}
\Crefname{algorithm}{Algorithm}{Algorithms}
\crefformat{algorithm}  {\cmlabRef{Algo.}{#1}{#2}{#3}}
\Crefformat{algorithm}  {\cmlabRef{Algorithm}{#1}{#2}{#3}}
\crefrangeformat{algorithm}%
  {{\color{cmlabPrefix}Algos.}~#3{\color{cmlabNumber}#1}#4--#5{\color{cmlabNumber}#2}#6}
\crefmultiformat{algorithm}%
  {\cmlabRef{Algos.}{#1}{#2}{#3}}%
  {, #2{\color{cmlabNumber}#1}#3}%
  {, #2{\color{cmlabNumber}#1}#3}%
  {, and #2{\color{cmlabNumber}#1}#3}

\usepackage[
  enable,
  applybold,
  labname={CMLab},
  leftlogo={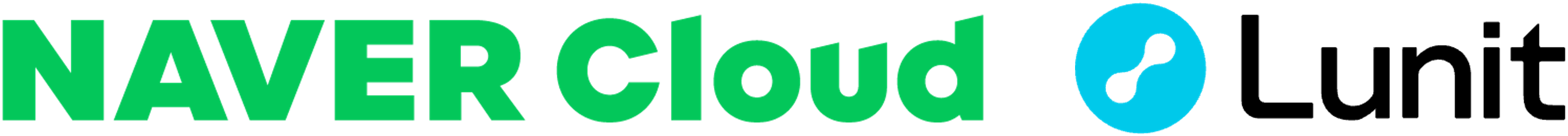},
  logo={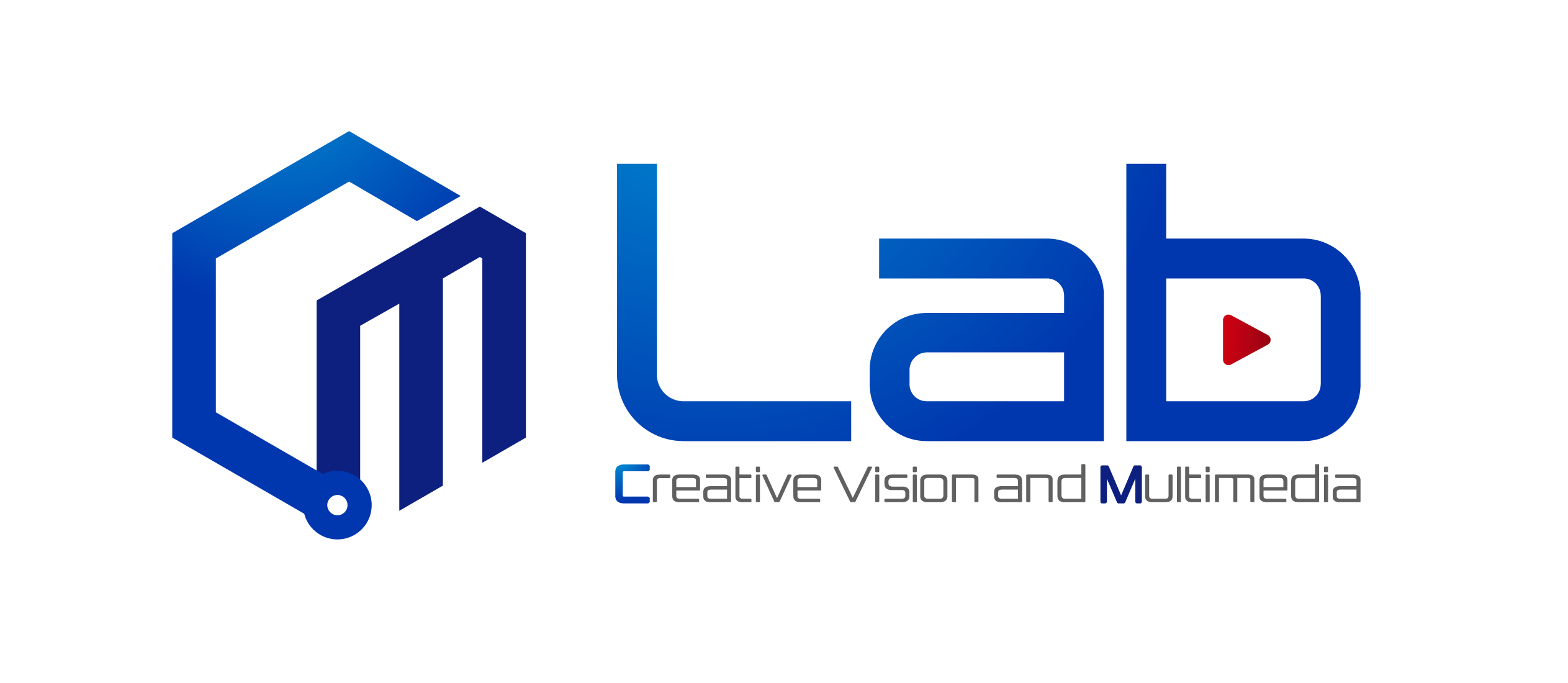},
  titleicon={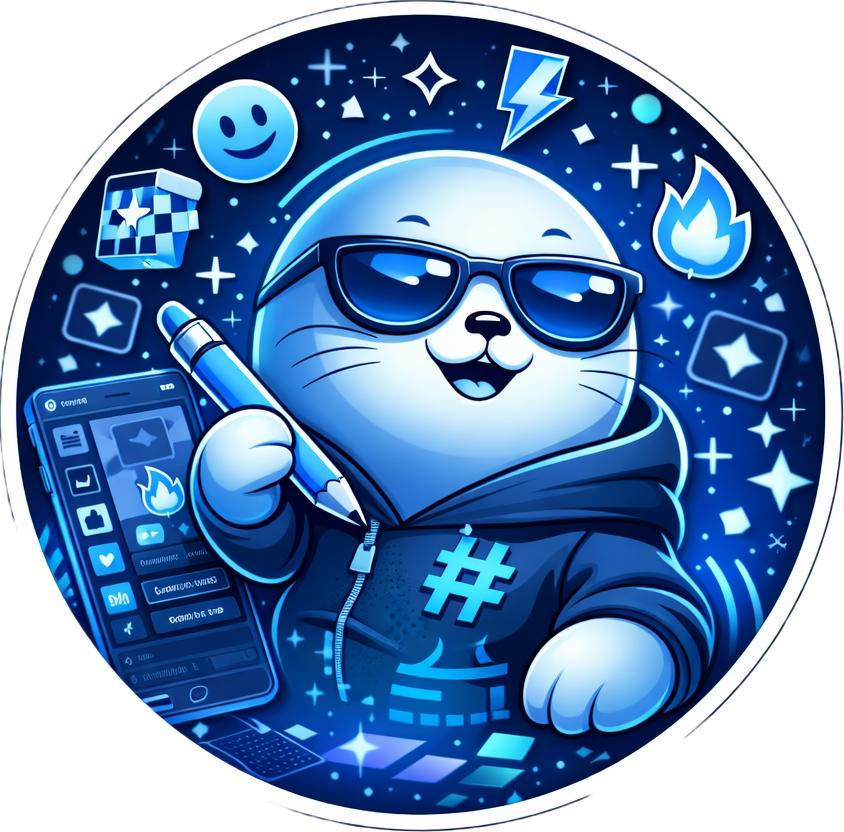},
  logowidth={22mm},
  projectpage={https://cmlab-korea.github.io/SEAL/}
]{cmlab_1}

\usepackage{booktabs} 
\usepackage{graphicx}
\usepackage{url}
\usepackage{xcolor}
\usepackage{placeins}
\usepackage{float}
\usepackage{pifont}

\graphicspath{{./figs/}}

\newcommand{\cmark}{\textcolor{green!60!black}{\ding{51}}}
\newcommand{\xmark}{\textcolor{red!70!black}{\ding{55}}}

\newcommand{\Ctoken}{C}         

\journal{Neurocomputing}

\begin{document}

\begin{frontmatter}

\title{SEAL: Semantic-aware Single-image\\
Sticker Personalization with a\\
Large-scale Sticker-tag Dataset}

\cmlabAuthors{Changhyun Roh$^{1}$ \qquad Yonghyun Jeong$^{2}$ \qquad Jonghyun Lee$^{3}$ \\ \qquad Chanho Eom$^{\dagger,1}$ \qquad Jihyong Oh$^{\dagger,1}$}
\cmlabAffiliations{$^{1}$Chung-Ang University \qquad $^{2}$NAVER Cloud \qquad $^{3}$Lunit Inc.}
\cmlabAuthorEmail{\{changhyunroh, cheom, jihyongoh\}@cau.ac.kr \quad yonghyun.jeong@navercorp.com \quad tomlee@lunit.io}
\cmlabProjectPage{https://cmlab-korea.github.io/SEAL/}

\begin{keyword}
Sticker Generation \sep Diffusion Model \sep Personalization
\end{keyword}

\end{frontmatter}

{
  \renewcommand{\thefootnote}{}
  \footnotetext{$^\dagger$ Co-corresponding authors.}
}

\section*{Abstract}
\label{sec:abstract}
Synthesizing a target concept from a single reference image remains challenging in diffusion-based personalized text-to-image generation, particularly in sticker personalization, where prompts frequently require explicit attribute edits. With only a single reference image, test-time fine-tuning (TTF) personalization methods often overfit to the reference. This overfitting typically appears as \textit{visual entanglement}, where background artifacts are absorbed into the learned concept representation, and \textit{structural rigidity}, where the model memorizes reference-specific spatial configurations and loses contextual controllability. To address these limitations, we introduce \textbf{SE}mantic-aware single-image sticker person\textbf{AL}ization (\textbf{SEAL}), a plug-and-play adaptation module that can be integrated into existing personalization pipelines without modifying their U-Net-based diffusion backbones. The SEAL module comprises three components applied during embedding adaptation: (1) a Semantic-guided Spatial Attention Loss, (2) a Split-merge Token Strategy, and (3) a Structure-aware Layer Restriction. To support sticker-domain personalization with attribute-level control, we introduce StickerBench, a large-scale sticker image dataset with structured tags under a six-attribute schema (Appearance, Emotion, Action, Camera Composition, Style, and Background). The resulting attribute-rich annotations provide a consistent interface for varying context while keeping the target identity fixed, enabling systematic evaluation of identity disentanglement and contextual controllability in single-image sticker personalization. Experiments show that integrating SEAL provides a more balanced trade-off between identity preservation and contextual controllability, highlighting the importance of explicit spatial and structural constraints during test-time adaptation. The code, structured annotations, data-processing pipeline, and project page will be publicly released.

\section{Introduction}
\label{sec:introduction}

Recent advances in diffusion-based generative models have improved text-to-image synthesis \citep{ho2020denoising, song2020denoising, rombach2022high, zhang2023adding}, notably in sticker personalization, where prompts frequently require explicit attribute edits. Personalization extends these models by adapting a user-provided concept to new textual contexts. Existing personalization approaches \citep{gal2022textual, ruiz2023dreambooth} can be broadly grouped into pre-trained adaptation models that generalize across instances \citep{shi2024instantbooth, zeng2024jedi, xie2025serialgen, wang2025ms} and test-time fine-tuning (TTF) methods that optimize model parameters or concept embeddings for a specific target \citep{gal2022textual, ruiz2023dreambooth, pang2024attndreambooth, wu2025core}. Test-time fine-tuning methods often provide strong identity preservation, but they become fragile when only a single reference image is available.

\begin{figure}[!t]
    \centering
    \includegraphics[width=0.95\columnwidth,height=0.33\textheight,keepaspectratio]{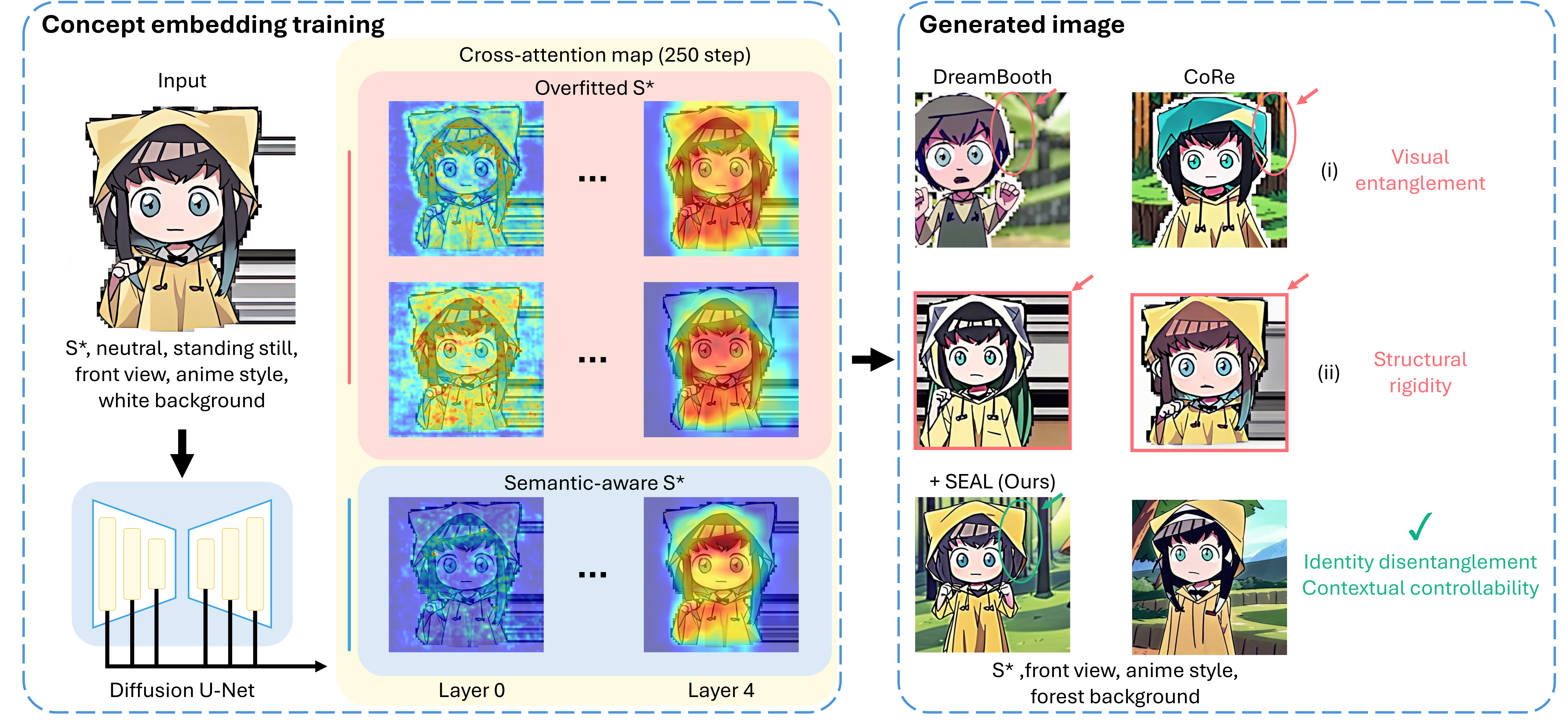}
    \caption{Comparison of cross-attention maps. We visualize spatial cross-attention maps of the concept token from an image-near cross-attention layer (Layer 0), which tends to emphasize low-level spatial patterns, and a semantically informative cross-attention layer (Layer 4). Baselines (top) such as DreamBooth \citep{ruiz2023dreambooth} and CoRe \citep{wu2025core} exhibit rigid patterns in Layer 0, leading to (i) \textit{visual entanglement} (red circle) and (ii) \textit{structural rigidity} (red rectangle). In contrast, our semantic-aware $S^*$ (bottom) suppresses these artifacts by enforcing object-aligned attention and reducing the influence of structurally biased image-near layers, thereby mitigating \textit{structural rigidity}, improving identity disentanglement from the background, and restoring contextual controllability under attribute-level prompt edits.} 
    \label{fig:teaser}
\end{figure}

TTF methods often achieve high-fidelity results when trained with multiple reference images, which provide cross-view variation that helps separate the target concept from its surrounding context. By observing the same instance under different backgrounds, poses, or compositions, the adaptation signal emphasizes identity-consistent cues while suppressing environment-specific artifacts. However, in sticker personalization, users often provide only a single reference image for convenience, making this multi-view assumption difficult to satisfy. In single-image settings, the absence of such variation makes it difficult to disentangle the target concept from the reference-specific context (\cref{fig:teaser}). Therefore, TTF personalization methods may absorb background patterns and fixed spatial configurations into the learned representation. This overfitting manifests as (i) \textit{visual entanglement}, where background artifacts are encoded together with concept identity, and (ii) \textit{structural rigidity}, where the model memorizes the layout of the reference image and loses contextual controllability.

These failure modes can be observed in cross-attention behavior. As shown in \cref{fig:teaser}, we visualize spatial cross-attention maps of the concept token extracted from the diffusion U-Net by aggregating attention weights over heads and reshaping them to the corresponding layer resolution. When a concept is adapted from a single image, these maps can leak into irrelevant background regions or align with rigid spatial patterns, indicating that the adaptation process couples semantic identity with the background and reference-specific layout.

These limitations are further amplified during test-time adaptation in the single-image setting, where the optimization signal can couple concept identity with background and reference-specific layout. Existing methods improve objective design or training procedures \citep{pang2024attndreambooth, wu2025core}, but they rarely impose explicit constraints that prevent background leakage, encourage attribute-diverse concept learning under one-shot supervision, and reduce reliance on structurally biased layers.

To address this gap, we introduce a semantic adaptation module that regulates embedding optimization for single-image sticker personalization without modifying the underlying diffusion architecture. The proposed module, consisting of three components, can be used as a plug-and-play method across U-Net-based diffusion architectures. First, we introduce a Semantic-guided Spatial Attention Loss that leverages object masks generated by the Segment Anything Model (SAM) \citep{kirillov2023segment}. This loss constrains the cross-attention maps of the concept token to align with the object region while suppressing activation in background areas, mitigating \textit{visual entanglement} during adaptation and improving identity disentanglement from the background. Second, a Split-merge Token Strategy optimizes auxiliary embeddings at different positions in the conditioning sequence, merges them by averaging at step $T_m$, and further optimizes the merged embedding during the remaining adaptation steps. The position-dependent optimization signals encourage attribute-diverse concept learning, improve optimization stability under extreme data scarcity, and help restore contextual controllability under attribute-level prompt edits. Third, we adopt a Structure-aware Layer Restriction that restricts the spatial constraint to semantically informative cross-attention layers, motivated by analyses of hierarchical representations in diffusion U-Nets \citep{zhu2025sdm}. By excluding shallow layers that encode low-level structural patterns, the adaptation process reduces \textit{structural rigidity} caused by layout memorization and restores contextual controllability for flexible generation.

To support controlled evaluation of single-image sticker personalization, we also introduce StickerBench, a large-scale sticker image dataset with structured tags under a six-attribute schema (Appearance, Emotion, Action, Camera Composition, Style, and Background). These attribute-rich annotations provide a consistent interface for varying context while keeping the target identity fixed, enabling systematic evaluation of \textit{identity disentanglement} and \textit{contextual controllability} under attribute-level prompt edits.

In summary, our contributions are as follows:
\begin{itemize}
    \item We introduce \textbf{SEAL}, a plug-and-play semantic adaptation module that can be integrated into existing TTF personalization pipelines without modifying their diffusion backbones.
    \item SEAL mitigates \textit{visual entanglement} via the \textit{Semantic-guided Spatial Attention Loss}, which aligns cross-attention maps of the concept token with the object region while suppressing background leakage; it alleviates \textit{structural rigidity} via the \textit{Split-merge Token Strategy} and \textit{Structure-aware Layer Restriction}, encouraging attribute-diverse concept learning and improving optimization stability under one-shot supervision.
    \item We introduce StickerBench, a large-scale sticker image dataset with structured tags under a six-attribute schema, enabling attribute-level prompt variations for evaluating identity disentanglement and contextual controllability in single-image sticker personalization.
    \item Experiments demonstrate that integrating SEAL into representative TTF personalization methods improves the balance between identity preservation and contextual controllability.
\end{itemize}  

\section{Related Work}
\label{sec:related_work}

\noindent
\paragraph{Text-to-Image Generative Models}
Recent advances in diffusion models \citep{ho2020denoising, song2020denoising, rombach2022high, zhang2023adding} have significantly improved text-to-image synthesis. Diffusion models generate high-fidelity images through a two-stage process: a forward diffusion process that progressively adds noise and a learnable reverse process that reconstructs the data. To improve efficiency, latent diffusion models (LDMs), such as Stable Diffusion \citep{rombach2022high}, perform diffusion in a compressed latent space. Conditional generation is commonly implemented via a cross-attention mechanism \citep{rombach2022high, zhang2023adding} within a U-Net backbone \citep{ronneberger2015u}, which mediates interactions between textual conditions and spatial visual features. More recent generators replace the U-Net with a diffusion transformer (DiT) backbone \citep{peebles2023scalable}. Multimodal DiT models, such as Stable Diffusion 3 \citep{esser2024scaling} and FLUX \citep{flux2024}, concatenate text and image tokens and process them with joint attention, so text-to-image interaction is no longer exposed as a dedicated cross-attention layer.

\noindent
\paragraph{Text-to-Image Personalization}
Text-to-image (T2I) personalization adapts pre-trained diffusion models to incorporate user-provided concepts. Existing approaches are broadly categorized into pre-trained adaptation (PTA) frameworks \citep{shi2024instantbooth, zeng2024jedi, xie2025serialgen, wang2025ms} and test-time fine-tuning (TTF) methods \citep{gal2022textual, ruiz2023dreambooth, wei2023elite, kumari2023multi, hao2023vico, huang2024classdiffusion, patel2024lambda, pang2024attndreambooth, wu2025core}. While TTF methods such as Textual Inversion and DreamBooth often provide strong identity preservation, they can severely overfit in data-scarce, one-shot (single-image) settings, where the model must infer the target concept from a single observation. In this regime, the adaptation signal can couple concept identity with background artifacts and reference-specific layout, leading to \textit{visual entanglement} and \textit{structural rigidity}. Recent studies have explored regularization strategies to alleviate these issues. For example, CoRe \citep{wu2025core} introduces Context Embedding Regularization to align the semantic context of the learned concept with its super-category, improving semantic consistency during adaptation. However, such context-level regularization does not directly constrain where the concept token attends in space and therefore may remain limited in preventing background leakage under one-shot supervision. Personalization has also been extended to DiT-based backbones. OminiControl \citep{tan2025ominicontrol} adapts diffusion transformers for subject-driven generation, and recent work explores segmentation-free multi-concept personalization \citep{phung2026universe}. These methods operate on backbones without dedicated cross-attention layers. Separately, T-LoRA \citep{soboleva2026t} targets overfitting in the single-image regime through rank-constrained adaptation, which shares our problem setting but not our attention-level formulation. Our approach is motivated by this gap and focuses on regulating the adaptation process itself, particularly for single-image settings in which explicit spatial and structural constraints are important for preserving identity while maintaining contextual controllability.

\noindent
\paragraph{Cross-Attention and Spatial Regularization}
Precise control of cross-attention is crucial for disentangling the subject from its context. Prior work has shown that erroneous attention maps can lead to failures in subject composition \citep{kumari2023multi}. To enforce boundary- and region-aware control, recent methods use mask-guided generation and attention regularization. Representative approaches constrain attention to meaningful regions (\textit{e.g.}, ViCo \citep{hao2023vico}) or introduce loss terms to suppress attention leakage. Such attention manipulation has recently been extended to multimodal DiT architectures, where text-to-image attention must be extracted from joint attention blocks rather than from dedicated cross-attention layers \citep{yin2025consistedit, wei2025freeflux}. Building on these principles but targeting the specific failure mode of one-shot \textit{visual entanglement}, our approach introduces a \textit{Semantic-guided Spatial Attention Loss}. We leverage SAM \citep{kirillov2023segment} to provide a spatial prior and use an attention-mask alignment term to guide concept-token cross-attention toward the object region while suppressing activation in background areas, thereby improving \textit{identity disentanglement}.

\section{Method}
\label{sec:method}

\begin{figure}[t]
    \centering
    \includegraphics[width=\columnwidth,height=0.5\textheight,keepaspectratio]{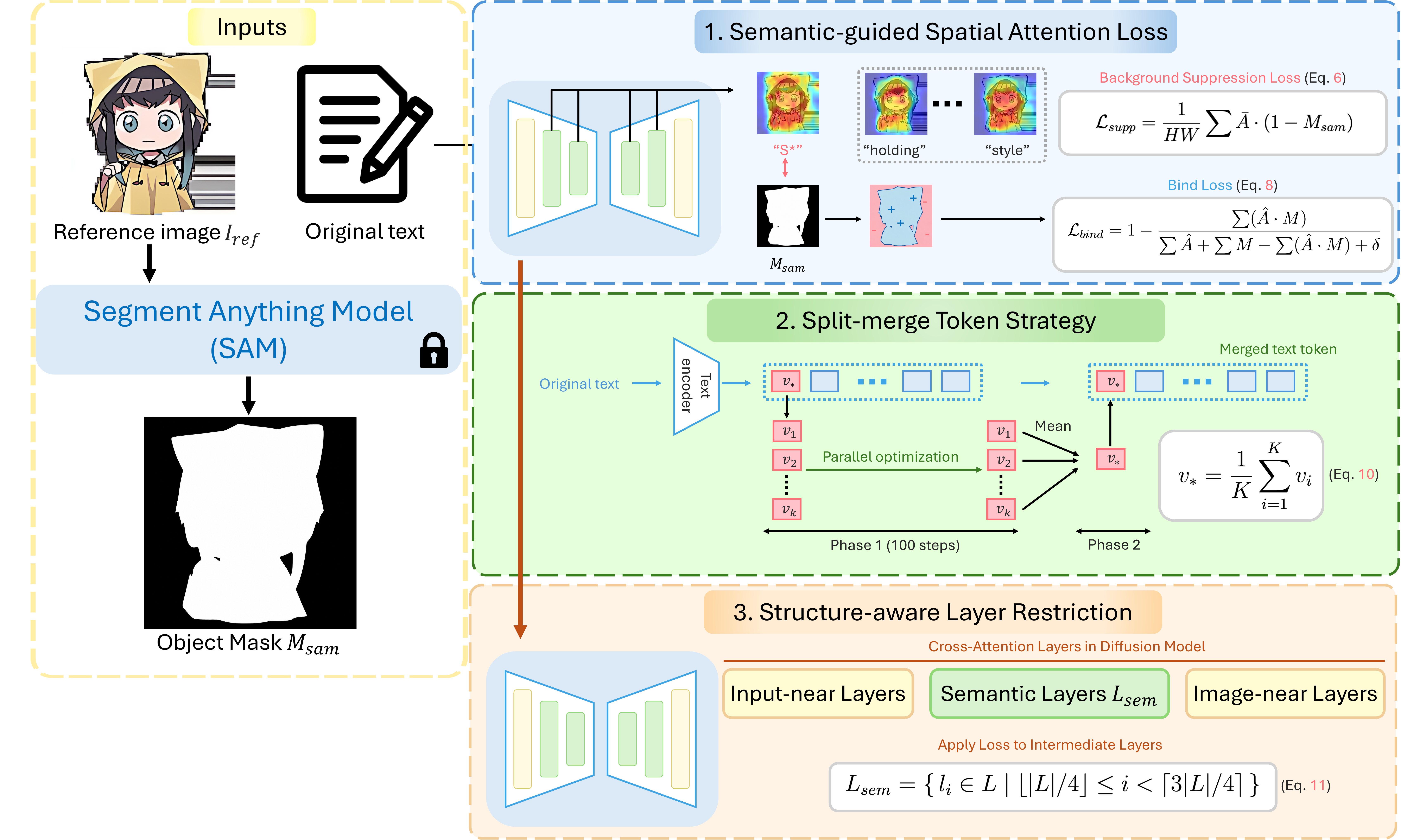}
    \caption{Overview of SEAL as a plug-and-play semantic adaptation module for single-image sticker personalization.}
    \label{fig:pipeline}
\end{figure}

This section presents \textbf{SEAL}, a plug-and-play adaptation module for single-image sticker personalization that can be integrated into existing personalization pipelines without modifying the diffusion backbone. \Cref{fig:pipeline} illustrates the overall architecture of the proposed method. We first introduce the diffusion formulation and notation used throughout the paper (\cref{subsec:prelim}). We then describe three components applied during embedding adaptation: (i) a Semantic-guided Spatial Attention Loss that constrains the concept token to attend to the object region to mitigate \textit{visual entanglement} (\cref{subsec:sam_loss}), (ii) a Split-merge Token Strategy that improves optimization stability under single-image supervision to alleviate \textit{structural rigidity} (\cref{subsec:split_merge}), and (iii) a Structure-aware Layer Restriction that applies the spatial constraint to semantically informative cross-attention layers to further reduce \textit{structural rigidity} and restore contextual controllability (\cref{subsec:layer_restriction}). Finally, we summarize the overall adaptation objective (\cref{subsec:overall_obj}).

\subsection{Preliminaries}
\label{subsec:prelim}

We consider single-image sticker personalization where a target concept is provided as a single reference image $I_{ref}$. Following TTF personalization settings \citep{gal2022textual, ruiz2023dreambooth, kumari2023multi}, we introduce a unique concept token $S^*$ into the conditioning text and optimize its concept embedding $v_*$.

We use a latent diffusion model \citep{rombach2022high}. Let $\mathcal{E}(\cdot)$ and $\mathcal{D}(\cdot)$ denote the variational autoencoder (VAE) encoder and decoder, respectively. We encode the reference image into a latent variable $z=\mathcal{E}(I_{ref})$. Let $t$ be a diffusion timestep sampled from a predefined schedule, and let $\alpha_t \in (0,1)$ be the corresponding noise schedule coefficient. We sample Gaussian noise $\epsilon \sim \mathcal{N}(0,I)$ and construct the noisy latent
\begin{equation}
\label{eq:noisy_latent}
z_t = \sqrt{\alpha_t}\, z + \sqrt{1-\alpha_t}\, \epsilon.
\end{equation}
Here, $z_t$ is the noisy latent at timestep $t$, and $I$ is the identity matrix. Let $\epsilon_\theta(\cdot)$ denote the U-Net noise predictor with parameters $\theta$. Let $c$ denote the text conditioning computed by a text encoder $\tau_\phi(\cdot)$ with parameters $\phi$. The conditioning $c$ contains the token $S^*$ whose embedding is the learnable vector $v_*$ in the text embedding space. The standard denoising objective is
\begin{equation}
\label{eq:diffusion_obj}
\mathcal{L}_{diffusion} = 
\mathbb{E}_{z,t,\epsilon,c}\left[\left\| \epsilon - \epsilon_\theta(z_t, t, c) \right\|_2^2 \right],
\end{equation}
where $\|\cdot\|_2$ is the Euclidean norm.

\paragraph{Cross-attention map for the concept token}
Let $l$ index a cross-attention layer in the U-Net. We denote by $A^{(l)}(S^*) \in \mathbb{R}^{H_l \times W_l}$ the spatial cross-attention map associated with token $S^*$ at layer $l$, obtained by selecting the attention weights for $S^*$, aggregating them over heads, and reshaping to the spatial resolution $(H_l, W_l)$ of that layer.

\subsection{Semantic-guided Spatial Attention Loss}
\label{subsec:sam_loss}

Single-image sticker personalization is underdetermined because the model observes only one instance of the concept with a fixed background and layout. During embedding adaptation, the concept token can absorb reference-specific context, causing background artifacts to be embedded together with identity cues. Since cross-attention determines where the concept token influences spatial features, constraining the spatial cross-attention maps of the concept token provides a direct way to mitigate \textit{visual entanglement}. We implement this constraint using an object mask predicted by SAM \citep{kirillov2023segment}.

Given the reference image $I_{ref}$, we obtain a mask $M_{sam}$ indicating the object region. Let $l$ index a cross-attention layer of the diffusion U-Net. We denote by $A \in \mathbb{R}^{H \times W}$ the spatial cross-attention map of the concept token $S^*$ at layer $l$, obtained by selecting the attention weights for $S^*$, aggregating them across heads, and reshaping them to the layer resolution $(H,W)$. We resize $M_{sam}$ to the same resolution and (if needed) binarize it by thresholding:
\begin{equation}
\label{eq:mask_bin}
M_{sam}=\mathbf{1}[M_{sam}>0.5], \qquad M_{sam}\in\{0,1\}^{H\times W}.
\end{equation}
A practical issue is that raw attention magnitudes vary across layers and timesteps, so the spatial constraint should be scale-invariant. We therefore apply $\ell_1$ normalization to the attention map over spatial locations:
\begin{equation}
\label{eq:attn_l1}
\bar{A}=\frac{A}{\sum A+\delta},
\end{equation}
where the summation is taken over all spatial locations and $\delta>0$ is a small constant for numerical stability. For the alignment term, we additionally sharpen the normalized map by squaring and renormalizing:
\begin{equation}
\label{eq:attn_sharpen}
\hat{A}=\frac{\bar{A}\odot\bar{A}}{\sum(\bar{A}\odot\bar{A})+\delta},
\end{equation}
where $\odot$ denotes element-wise multiplication. This sharpening emphasizes consistently activated regions and reduces the effect of diffuse attention, making the alignment constraint more sensitive to the object boundary.

We impose two complementary constraints. The first term suppresses attention leakage into background regions by penalizing attention mass outside the object mask:
\begin{equation}
\label{eq:supp}
\mathcal{L}_{supp}=\frac{1}{HW}\sum\left(\bar{A}\odot(1-M_{sam})\right).
\end{equation}
This term discourages the concept token from attending to background textures and surrounding regions that co-occur with the concept in $I_{ref}$.

The second term measures the alignment between the attention support and the object region. We normalize the binary mask over spatial locations:
\begin{equation}
\label{eq:mask_norm}
M=\frac{M_{sam}}{\sum M_{sam}+\delta}.
\end{equation}
We then compute
\begin{equation}
\label{eq:bind}
\mathcal{L}_{bind}=1-\frac{\sum(\hat{A}\odot M)}{\sum \hat{A}+\sum M-\sum(\hat{A}\odot M)+\delta}.
\end{equation}
The fraction in \cref{eq:bind} is a monotonic transformation of the dot-product affinity between the normalized attention map and mask. We use $\mathcal{L}_{bind}$ as an attention-mask alignment regularizer during adaptation. The spatial attention constraint at layer $l$ is defined as:
\begin{equation}
\label{eq:spatial_layer}
\mathcal{L}_{spatial}^{(l)}=\lambda_{bind}\mathcal{L}_{bind}+\lambda_{supp}\mathcal{L}_{supp},
\end{equation}
where $\lambda_{bind}$ and $\lambda_{supp}$ control the contributions of alignment and suppression. In the overall objective (\cref{subsec:overall_obj}), we compute $\mathcal{L}_{spatial}^{(l)}$ over selected layers and aggregate them into a single regularizer. This design applies spatial supervision directly to the concept token during embedding adaptation and does not require architectural changes to the diffusion backbone.

\subsection{Split-merge Token Strategy}
\label{subsec:split_merge}

Optimizing a single concept embedding from a single reference image is often unstable \citep{park2024textboost}, and the learned representation may capture only a narrow subset of the concept attributes. To stabilize embedding adaptation and encourage attribute-diverse concept learning, we introduce a Split-merge Token Strategy. Instead of directly optimizing one vector for the concept token $S^*$, we instantiate a set of $K$ auxiliary embeddings $\mathcal{V}_{aux}=\{v_1,\dots,v_K\}$, all initialized from the embedding of the super-category token $\Ctoken$. Although initialized from the same super-category embedding, the auxiliary tokens occupy different positions in the conditioning sequence and therefore receive position-dependent optimization signals. During the first $T_m$ adaptation steps, the $K$ embeddings are optimized in parallel. At step $T_m$, the auxiliary embeddings are merged by averaging:
\begin{equation}
\label{eq:splitmerge}
v_* = \frac{1}{K}\sum_{i=1}^{K} v_i.
\end{equation}
The merged embedding $v_*$ is further optimized during the remaining adaptation steps and is used to condition the diffusion model at inference. By aggregating the position-dependent optimization outcomes, the strategy encourages the learned representation to retain a broader range of concept attributes.

\subsection{Structure-aware Layer Restriction}
\label{subsec:layer_restriction}

The choice of which cross-attention layers receive spatial supervision and diffusion-model updates determines what information is learned by the concept embedding. Cross-attention in diffusion U-Nets is hierarchical: earlier layers are closer to the input representation and tend to emphasize low-level spatial patterns, while deeper layers increasingly encode semantic information \citep{zhu2025sdm, kye2025chimera}. In one-shot personalization, this hierarchy becomes critical. If the spatial constraint is enforced on layers that are close to the input representation, the concept embedding can become tied to edges, textures, and the fixed layout of the reference image. This encourages memorization and amplifies structural rigidity. Conversely, enforcing the constraint in more semantically oriented layers promotes learning identity-relevant information while reducing sensitivity to the exact spatial configuration of the reference.

A common implementation choice is to select layers based on a resolution-specific notion of a bottleneck. However, resolution alone is not a reliable proxy for semantic depth across diffusion backbones. In practice, multiple cross-attention blocks can appear at the same spatial resolution while occupying different depths and playing different roles in adaptation. As a result, a rule that depends on identifying a particular resolution stage can be fragile when transferring across architectures. To preserve compatibility, we define layer restriction by relative depth in the ordered sequence of cross-attention layers, rather than by a resolution-based bottleneck.

Let $L=\{l_0,\dots,l_{|L|-1}\}$ denote the ordered set of cross-attention layers from which the concept-token cross-attention map $\hat{A}$ can be extracted, and let $|L|$ be the number of such layers. We select semantically informative layers by relative depth and use a fixed central range of the ordered sequence:
\begin{equation}
\label{eq:lsem}
L_{sem} = \{\, l_i \in L \mid \lfloor |L|/4 \rfloor \le i < \lceil 3|L|/4 \rceil \,\}.
\end{equation}
In practice, Structure-aware Layer Restriction is implemented as a layer-wise optimization mask over the diffusion adaptation path: layers in $L_{sem}$ retain the learning rate specified by the underlying personalization method, whereas the learning rates of the remaining cross-attention layers are set to zero. Therefore, the layer restriction can be applied independently even when the spatial loss is disabled.
This choice avoids backbone-specific hyperparameters and does not rely on identifying resolution stages, which can vary across diffusion backbones. The motivation is that cross-attention representations become increasingly mixed and abstract with depth: enforcing spatial constraints on input-near layers can amplify \textit{structural rigidity} by tying the embedding to local patterns and reference-specific structure, while layers close to output reconstruction can be biased toward reproducing fine spatial details. Selecting a fixed central range provides a simple, backbone-independent way to emphasize semantically oriented supervision while de-emphasizing input-near and output-near structural biases. We adopt this fixed central-range rule as a deterministic design choice (rather than a tunable setting) and apply it uniformly across all experiments.

During embedding adaptation, we compute the Semantic-guided Spatial Attention Loss (\cref{subsec:sam_loss}) only on layers in $L_{sem}$ and aggregate it by averaging:
\begin{equation}
\label{eq:spatial_agg}
\mathcal{L}_{spatial} = \frac{1}{|L_{sem}|}\sum_{l \in L_{sem}} \mathcal{L}_{spatial}^{(l)}.
\end{equation}
Here, $|L_{sem}|$ is the number of selected layers and $\mathcal{L}_{spatial}^{(l)}$ is defined in \cref{eq:spatial_layer}. We use averaging so that the aggregated spatial constraint reflects a stable supervision signal across the selected semantic layers while reducing sensitivity to layer-specific outlier responses. This is particularly important in single-image sticker personalization, where attention behavior can vary noticeably across layers and a specific layer may over-emphasize only a partial region or reference-specific local pattern of the target concept. Averaging mitigates such layer-specific bias and applies the spatial prior in a distributed manner across the selected semantic layers, encouraging supervision over the concept as a whole rather than over a small subset of its regions. By emphasizing the central, semantically oriented layers and de-emphasizing structurally biased layers, this strategy reduces \textit{structural rigidity} while preserving identity-relevant supervision.

\subsection{Overall Adaptation Objective}
\label{subsec:overall_obj}

We optimize the concept embedding associated with token $S^*$ for single-image sticker personalization while keeping the diffusion backbone architecture unchanged. The adaptation objective combines the standard diffusion denoising loss with the proposed Semantic-guided Spatial Attention Loss. The denoising term preserves the base diffusion training signal used in TTF personalization \citep{gal2022textual, ruiz2023dreambooth, kumari2023multi}:
\begin{equation}
\label{eq:diffusion_obj_repeat}
\mathcal{L}_{diffusion} = 
\mathbb{E}_{z,t,\epsilon,c}\left[\left\| \epsilon - \epsilon_\theta(z_t, t, c) \right\|_2^2 \right],
\end{equation}
where $z_t$ is constructed from the reference latent $z=\mathcal{E}(I_{ref})$ and Gaussian noise $\epsilon \sim \mathcal{N}(0,I)$ (\cref{eq:noisy_latent}), and $c$ is the text conditioning that includes the concept token $S^*$ with its learnable embedding.

The spatial regularizer is computed using the SAM mask $M_{sam}$ \citep{kirillov2023segment} and the cross-attention maps of $S^*$ extracted from selected layers. Concretely, for each layer $l \in L_{sem}$ (\cref{eq:lsem}), we compute the layer-wise spatial loss $\mathcal{L}_{spatial}^{(l)}$ (\cref{eq:spatial_layer}) using the normalized attention map $\hat{A}^{(l)}(S^*)$. We then aggregate the spatial constraint by averaging over the selected layers:
\begin{equation}
\label{eq:spatial_agg_repeat}
\mathcal{L}_{spatial} = \frac{1}{|L_{sem}|}\sum_{l \in L_{sem}} \mathcal{L}_{spatial}^{(l)}.
\end{equation}
This aggregation avoids reliance on a single layer and applies supervision at semantically informative depths as described in \cref{subsec:layer_restriction}.

The Split-merge Token Strategy (\cref{subsec:split_merge}) is integrated at the embedding level. During adaptation, we optimize the auxiliary embeddings $\{v_1,\dots,v_K\}$ in parallel and form the merged embedding $v_*$ via \cref{eq:splitmerge}. The spatial constraint is computed consistently with the current concept embedding used in conditioning, and the final merged embedding is used as the concept representation after aggregation.

The overall adaptation objective is defined as
\begin{equation}
\label{eq:total}
\mathcal{L}_{total} = \mathcal{L}_{diffusion} + \lambda_{spatial}\mathcal{L}_{spatial},
\end{equation}
where $\lambda_{spatial}$ controls the strength of the spatial constraint relative to the diffusion objective. Since the module only requires access to the concept embedding and the corresponding cross-attention maps, it can be integrated as a plug-and-play module into existing TTF adaptation procedures without architectural modifications to the diffusion backbone.
\begin{figure*}[t]
    \centering
    \includegraphics[width=\textwidth]{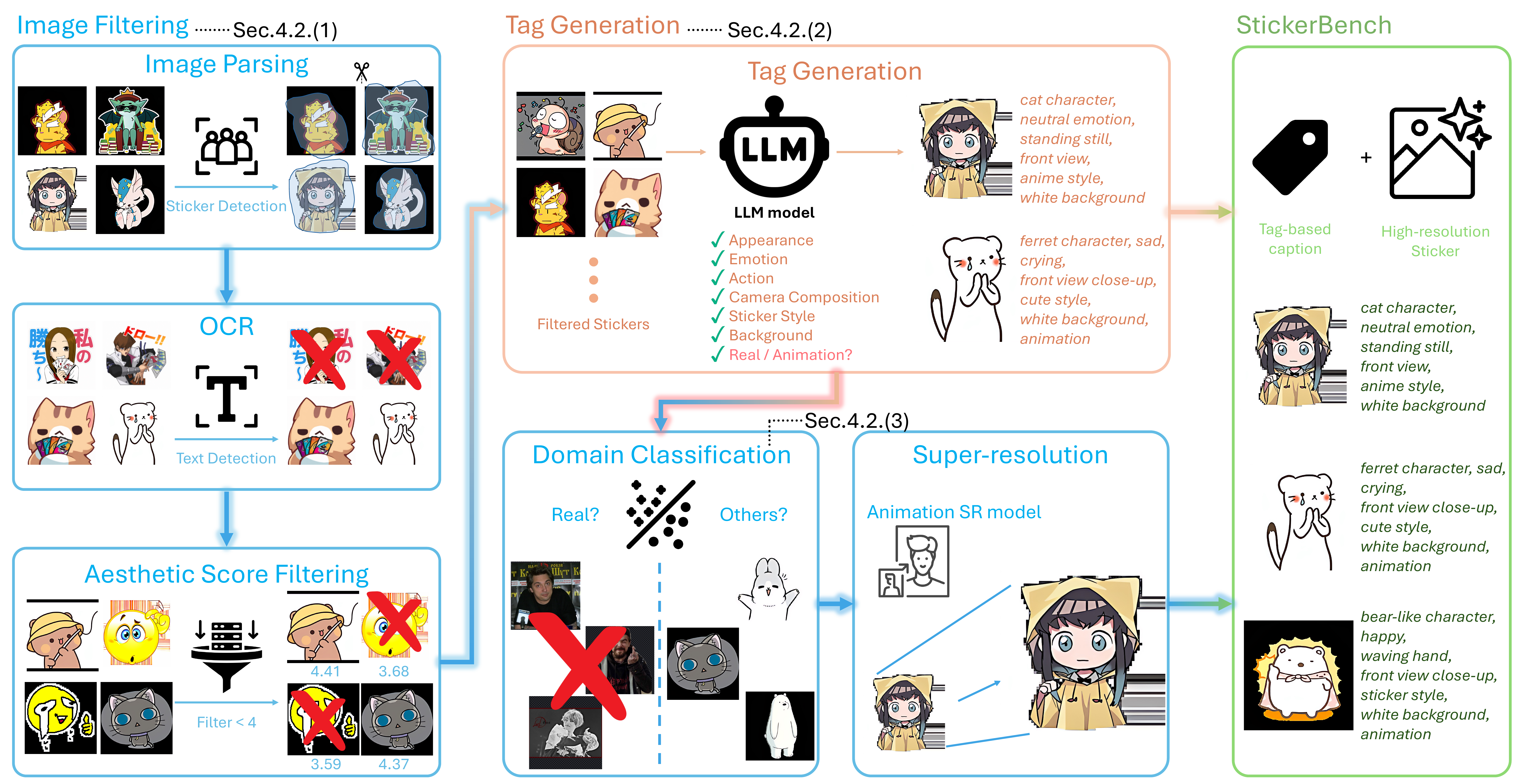}
    \caption{Overview of the proposed dataset construction pipeline. The framework comprises image parsing and filtering, tag generation, domain classification, and super-resolution, resulting in the curated StickerBench dataset.}
    \label{fig:framework_overview}
\end{figure*}
\section{StickerBench Dataset}
\label{sec:dataset_pipeline}
Sticker personalization frequently requires explicit attribute edits (\textit{e.g.}, changing the background, action, or composition) while keeping the target identity fixed. This makes single-image sticker personalization particularly sensitive to two dominant failure modes: (i) \textit{visual entanglement} and (ii) \textit{structural rigidity}. To support systematic evaluation of identity disentanglement and contextual controllability under such attribute-level edits, we introduce StickerBench, a large-scale sticker dataset paired with structured tag annotations.

\begin{figure*}[t]
    \centering
    \includegraphics[width=0.8\textwidth]{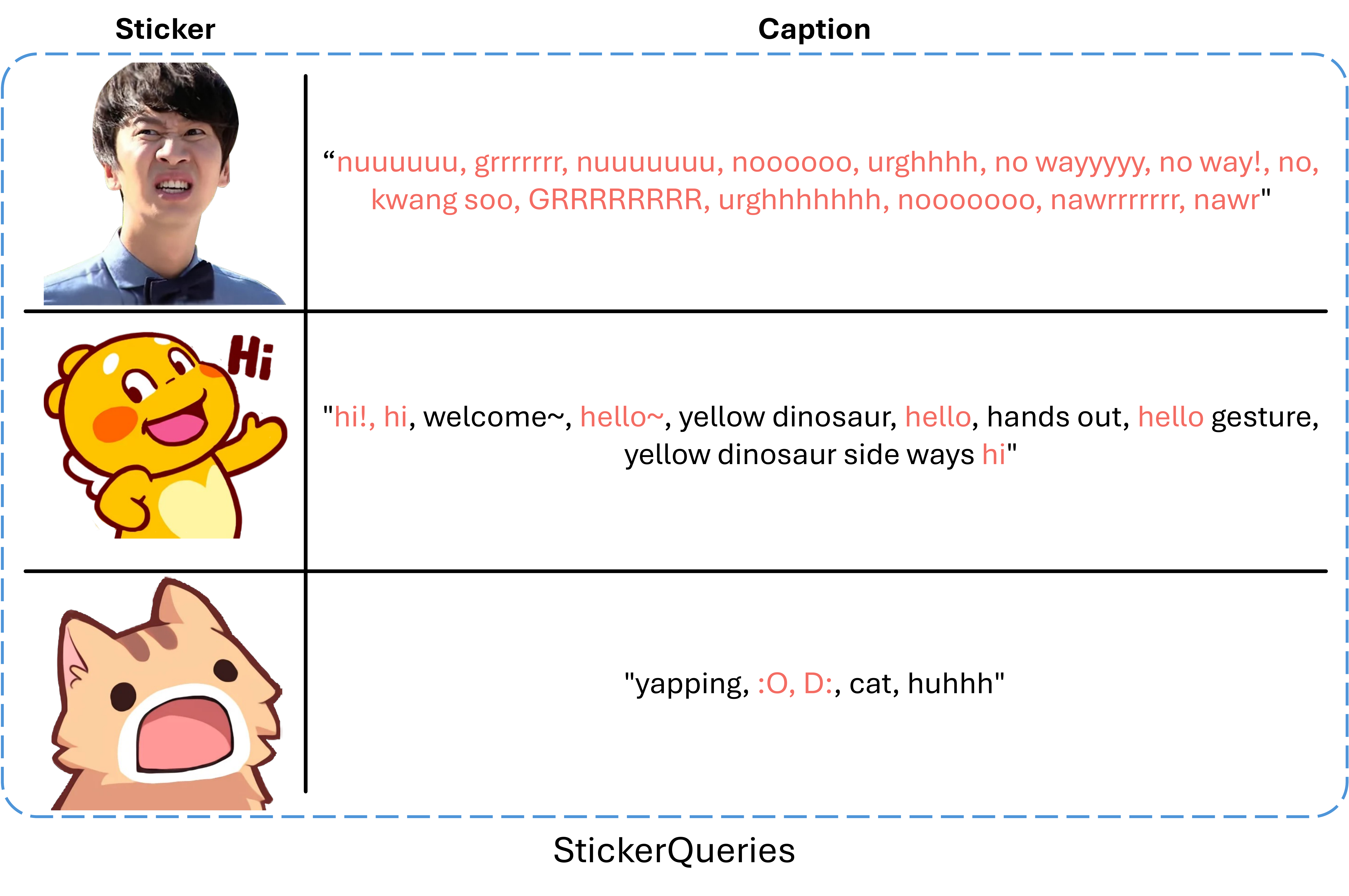}
    \caption{Examples from StickerQueries \citep{chee2025small}. The associated text annotations are short, informal, and highly variable in form, often consisting of exclamations, repeated words, or emoticons rather than structured semantic descriptions. Such annotations provide limited support for attribute-level prompt construction and controlled evaluation in single-image sticker personalization.}
    \label{fig:caption_examples}
\end{figure*}
\subsection{Motivation for a Tag-based Schema}
\label{subsec:tag_motivation}

Existing sticker datasets \citep{fei2021towards, liu2022ser30k, shi2024integrating, chee2025u, chee2025small} are primarily curated for retrieval or recognition and provide limited supervision for controllable generation. In particular, sentence-like captions are often noisy and inconsistent, which makes it difficult to isolate and manipulate specific semantic factors during evaluation, as illustrated in \cref{fig:caption_examples}. StickerBench is designed to address this gap by providing an attribute-factorized tag schema that supports controlled prompt edits and analysis in the sticker domain. A central requirement in single-image sticker personalization is to synthesize the adapted concept under diverse conditions while preserving identity. In the sticker domain, this requirement is more explicit because user prompts often request discrete changes in appearance, emotion, action, composition, style, and background. Evaluating such controllability benefits from a representation that separates these factors. A common approach is to describe each image using a sentence caption. However, sentence-based captions are free-form and frequently mix multiple factors into a single phrase, making it unclear which words correspond to which visual attributes. They can also contain semantic redundancy or phrasing variation that is unrelated to the controllability factors of interest. This mismatch complicates controlled prompt edits and makes it harder to diagnose whether failures are due to \textit{visual entanglement} (background cues carried over with identity) or \textit{structural rigidity} (layout memorization despite context changes).

\begin{table}[t]
\centering
\caption{LLaVA-based prompting protocol for structured tag annotation in StickerBench. The first field \texttt{<domain>} (\texttt{animation} or \texttt{real}) is used only for domain filtering and is excluded from the six-attribute schema. We map \texttt{<character>} to Appearance and \texttt{<composition>} to Camera Composition.}
\label{tab:tag_prompt_main}
\scriptsize
\renewcommand{\arraystretch}{0.8}
\begin{tabular}{p{0.85\columnwidth}}
\toprule
\textbf{System instruction (summary)} \\
Generate exactly one line with 7 comma-separated fields: \\
\texttt{<domain>, <character>, <emotion>, <action>, <composition>, <style>, <background>} \\
\texttt{<domain>} $\in \{\texttt{animation}, \texttt{real}\}$. \\
\midrule
\textbf{Constraints and formatting rules} \\
(1) Provide one primary tag per field. \\
(2) Use concise, standardized terms; avoid subjective phrasing. \\
(3) Describe only observable visual content. \\
(4) If a field is not applicable, output \texttt{none} or \texttt{not applicable}. \\
(5) Output must be strictly comma-separated with no additional text. \\
\bottomrule
\end{tabular}
\end{table}

To address these limitations, we adopt a tag-based representation that decomposes descriptions into attribute-level labels. StickerBench uses a six-attribute schema consisting of \textit{Appearance, Emotion, Action, Camera Composition, Style, and Background}. Our six-attribute schema is designed around the controllable factors that are most relevant to single-image sticker personalization. Appearance serves as the primary identity anchor, while Emotion and Action capture expression and pose variations that are frequently requested in sticker prompts. Camera Composition represents view and framing changes that are closely related to structural rigidity, and Style and Background capture rendering and contextual variations that are important for evaluating disentanglement from reference-specific cues. We do not assume that these attributes are perfectly independent in a semantic sense; rather, they provide a practical and consistent factorization for prompt construction and controlled evaluation in the sticker domain. Each sticker is annotated with tags for these attributes, enabling structured prompt construction and attribute-level evaluation by replacing tags for a target attribute while keeping other fields unchanged. This provides a consistent interface to vary context while holding identity cues fixed, supporting evaluation of \textit{identity disentanglement} and \textit{contextual controllability} in single-image sticker personalization settings.

\begin{table}[t]
    \centering
    \caption{Comparison of StickerBench with prior datasets. StickerBench provides a larger scale and structured annotations across six attributes.}
    \label{tab:dataset_comparison}
    \resizebox{0.90\columnwidth}{!}{
    \begin{tabular}{l r c c c c c c}
        \toprule
         \textbf{Dataset} & \textbf{Size} & \textbf{App.} & \textbf{Emo.} & \textbf{Act.} & \textbf{Comp.} & \textbf{Sty.} & \textbf{Bg.} \\
        \midrule
        MOD \citep{fei2021towards} & 0.3k & \xmark & \xmark & \xmark & \xmark & \xmark & \xmark \\
        SER30K \citep{liu2022ser30k} & 30k & \xmark & \cmark & \xmark & \xmark & \xmark & \xmark \\
        MCDSCS \citep{shi2024integrating} & 14k & \xmark & \xmark & \xmark & \xmark & \xmark & \xmark \\
        StickerQueries \citep{chee2025small} & 1k & \xmark & \cmark & \xmark & \xmark & \xmark & \xmark \\
        \midrule
        \textbf{Ours} & \textbf{261k} & \cmark & \cmark & \cmark & \cmark & \cmark & \cmark \\
        \bottomrule
    \end{tabular}%
    }

    \vspace{2pt}
    {\scriptsize{App.=Appearance, Emo.=Emotion, Act.=Action, Comp.=Camera Composition, Sty.=Style, Bg.=Background.}}
\end{table}
\subsection{Dataset Construction}
\label{subsec_pm:construction}

To construct StickerBench, we first aggregate an unfiltered collection of 646,508 images from public platforms, including Anita~\citep{cite_anita} and Telegram stickers~\citep{cite_telegram}. From this pool, we curate a final dataset of 261k stickers through a pipeline designed to improve semantic precision and stylistic consistency, as shown in \cref{fig:framework_overview}. The process is organized into three stages:

\paragraph{(1) Image Filtering}
We employ SAM \citep{kirillov2023segment} for instance extraction to segment individual sticker regions. We remove text-heavy and low-quality samples using an optical character recognition (OCR) model \citep{huang2024bridging} and an aesthetic score predictor \citep{aesthetic_predictor}, respectively.

\paragraph{(2) Tag Generation}
We generate structured tag annotations using a multimodal large language model (LLaVA-13B) \citep{liu2023visual} and organize the outputs into the six-attribute schema. This representation supports controlled prompt construction and analysis by separating semantic factors into fixed fields. To maintain a consistent annotation structure at scale, we use a deterministic system instruction that forces the output to follow a fixed schema with exactly seven comma-separated fields:
\texttt{<domain>}, \texttt{<character>}, \texttt{<emotion>}, \texttt{<action>},
\texttt{<composition>}, \texttt{<style>}, and \texttt{<background>}. The \texttt{<character>} field denotes the concept category in a class-like form (\textit{e.g.}, \texttt{bear character}). We impose constraints to encourage one primary tag per field, concise and standardized wording, and descriptions grounded in observable visual content. We summarize the protocol in \cref{tab:tag_prompt_main}. The first \texttt{<domain>} field (\texttt{animation}/\texttt{real}) is used only for domain classification and filtering and is discarded after curation.
The remaining six fields are used as the six-attribute schema, where \texttt{<character>} corresponds to Appearance and \texttt{<composition>} corresponds to Camera Composition. 
\begin{figure}[t]
    \centering
    \includegraphics[width=0.7\columnwidth]{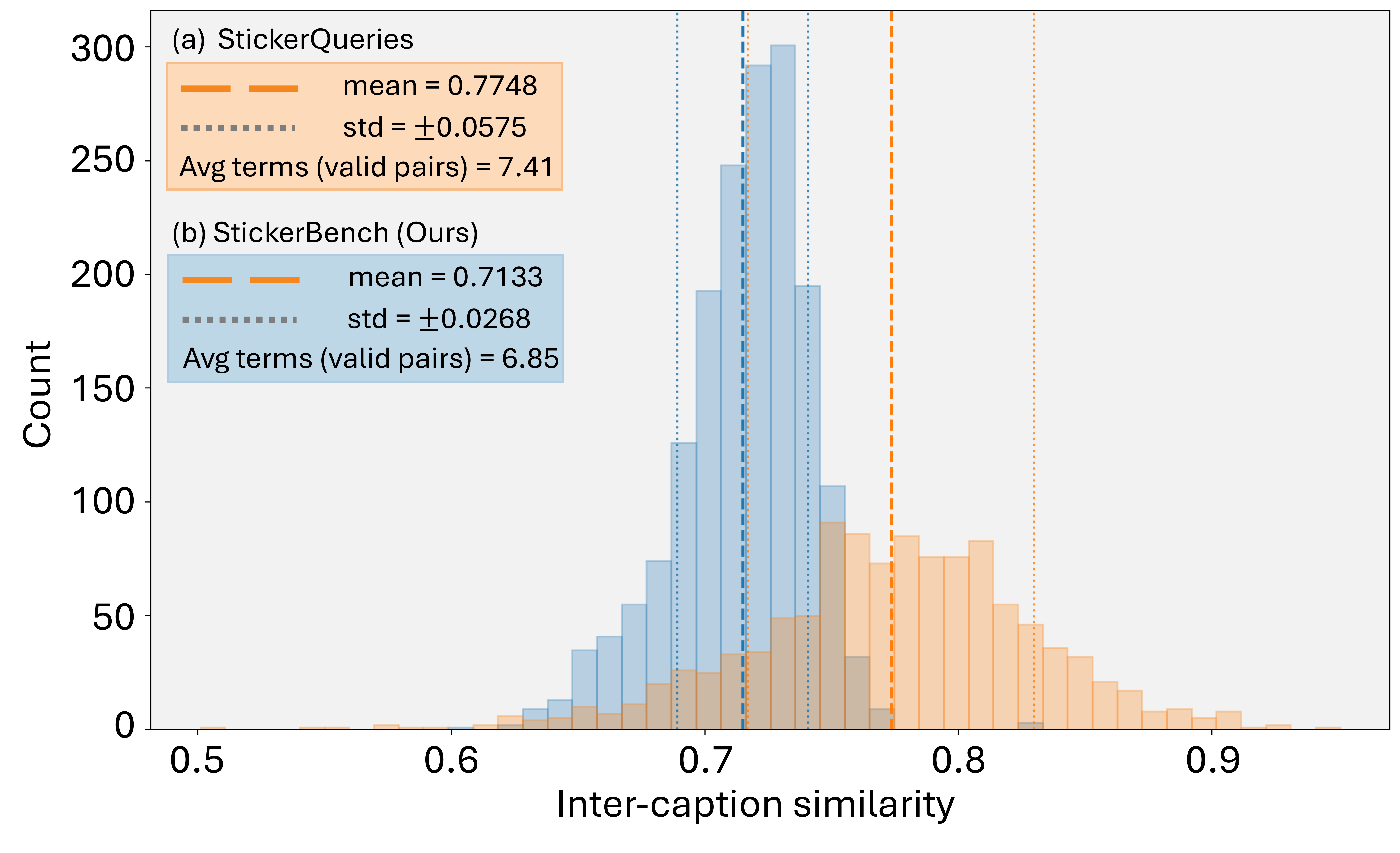}
    \caption{Comparison of intra-caption tag-similarity histograms based on CLIP text embeddings. (a) StickerQueries exhibits a mean cosine similarity of 0.77, indicating greater semantic redundancy within captions. (b) StickerBench shows a lower mean similarity of 0.71, suggesting that its six-attribute schema reduces redundancy among tags.}
    \label{fig:cosine_similarity}
\end{figure}
\paragraph{(3) Domain Classification and Alignment}
To ensure stylistic consistency, we perform domain classification to filter out photorealistic images. After the prior filtering steps, many remaining real-domain samples are raw photographs (\textit{e.g.}, screenshots or unrelated photography) rather than valid sticker instances. We exclude these images to reduce non-sticker noise. As a final step, we apply an animation-oriented super-resolution model \citep{wang2024apisr} to mitigate quality degradation.

\subsection{Dataset Analysis}
\label{subsec_pm:validation}
\begin{figure}[t]
    \centering
    \includegraphics[width=0.9\columnwidth]{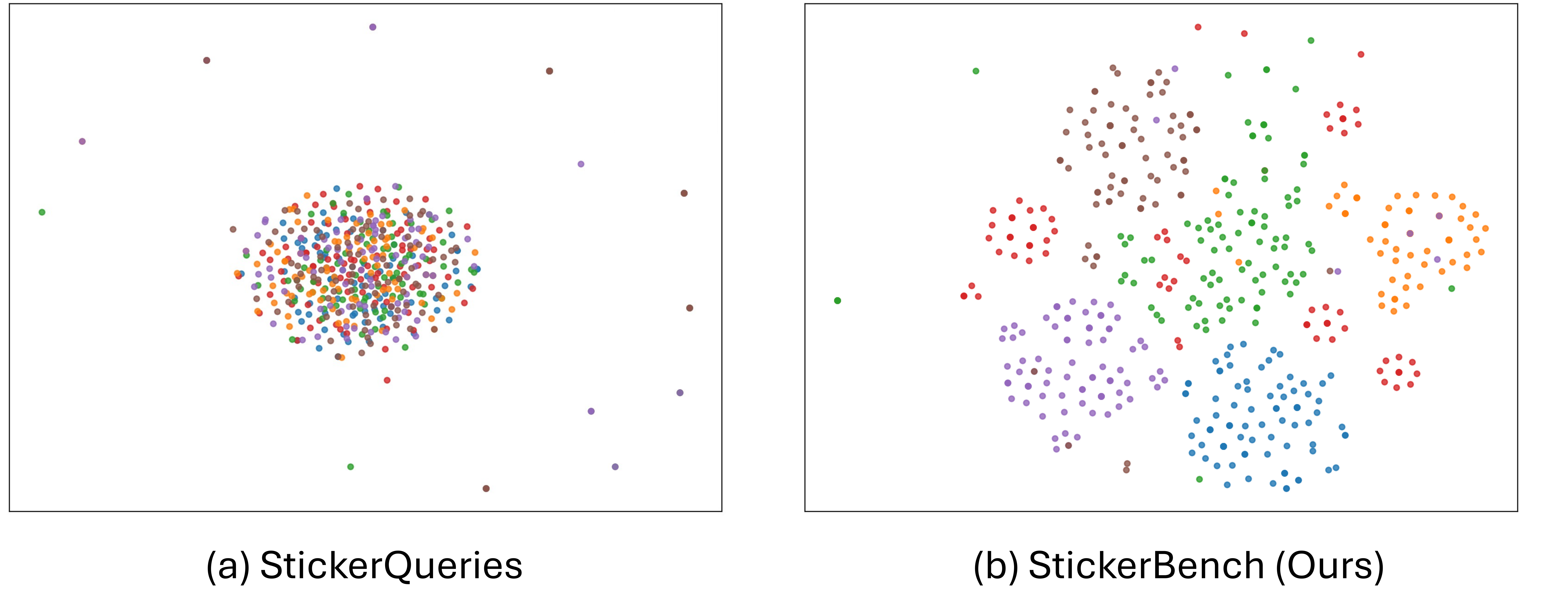}
    \caption{t-SNE visualization of tag embeddings. (a) Tags from existing datasets form a denser distribution with substantial overlap. (b) StickerBench tags form more distinct clusters aligned with the six attributes in our schema (Appearance, Emotion, Action, Camera Composition, Style, and Background). The visualization suggests that the structured schema separates the attribute categories more clearly than the sentence-like annotations in the comparison dataset.}
    \label{fig:tsne}
\end{figure}
We compare StickerBench with prior datasets in \cref{tab:dataset_comparison}. StickerBench provides a substantially larger scale and attribute-level structure across all six factors. To quantify redundancy in text representations, we compute intra-caption similarity using CLIP text embeddings \citep{radford2021learning}. As shown in \cref{fig:cosine_similarity}, StickerBench exhibits lower mean similarity than sentence-like captions, indicating reduced redundancy for attribute-level prompt construction. We further visualize tag embeddings using t-distributed stochastic neighbor embedding (t-SNE). \Cref{fig:tsne} shows that tags form structured clusters aligned with the six attributes, suggesting that the schema organizes semantic factors in a way that supports controlled analysis for personalization.

Finally, we emphasize that the proposed adaptation module (\cref{sec:method}) is independent of the annotation format. StickerBench is introduced to provide an attribute-factorized prompt interface for controlled evaluation of single-image sticker personalization in the sticker domain.

\paragraph{Human Validation of StickerBench Annotations}
We recruited 20 participants to evaluate the structured annotations of StickerBench. Ten participants reported being familiar with or working in AI/ML, while the other ten reported limited familiarity with the field. Each participant was shown 30 reference images sampled from the dataset together with their assigned tags. For each image, participants independently rated how well each tag described the visible image content on a scale from 0 to 3. A score of 0 indicates no correspondence, while a score of 3 indicates that the tag describes the image very well.

Participation was voluntary. Before participation, all participants were informed of the study purpose and that their responses would be collected anonymously and used for research, and they provided informed consent.

For each participant and attribute category, we first average the ratings over the 30 reference images. We then compute the Mean Opinion Score (MOS) and standard deviation across participants. For the Overall score, we first average the six attribute scores for each participant and then compute the mean and standard deviation across participants.

\begin{table}[t]
    \centering
    \caption{Human validation of StickerBench annotations by 20 participants. Participants rated the correspondence between each tag and the visible content of the reference image on a scale from 0 to 3. We report the Mean Opinion Score (MOS) and standard deviation across participants. Higher values indicate stronger correspondence.}
    \label{tab:annotation_audit}
    \resizebox{0.95\textwidth}{!}{%
        \begin{tabular}{l ccccccc}
            \toprule
            \textbf{Metric}
                & \textbf{Appearance}
                & \textbf{Emotion}
                & \textbf{Action}
                & \textbf{Camera Composition}
                & \textbf{Style}
                & \textbf{Background}
                & \textbf{Overall} \\
            \midrule
            \textbf{MOS} $\uparrow$
                & $2.86 \pm 0.19$
                & $2.44 \pm 0.41$
                & $2.47 \pm 0.39$
                & $2.38 \pm 0.56$
                & $2.73 \pm 0.24$
                & $2.37 \pm 0.47$
                & $2.54 \pm 0.34$ \\
            \bottomrule
        \end{tabular}%
    }
\end{table}

\Cref{tab:annotation_audit} shows generally strong correspondence between the structured tags and the visible image content. Appearance and Style receive the highest mean ratings, while Camera Composition and Background exhibit greater participant-level variation. The overall MOS of $2.54 \pm 0.34$ on the 0--3 scale supports the use of the structured annotations for constructing attribute-level prompts in StickerBench.

\section{Experiments}
\label{sec:experiments}

In this section, we provide the experimental settings, including implementation details and evaluation metrics (\cref{subsec:settings}). We then evaluate the transferability of the proposed adaptation module by integrating it into representative personalization methods and reporting quantitative and qualitative results (\cref{subsec:comparisons}). Our evaluation focuses on whether the module mitigates \textit{visual entanglement} and \textit{structural rigidity}, thereby improving identity disentanglement and contextual controllability in single-image sticker personalization. Finally, we provide ablation studies and diagnostic analyses to validate the role of each module component (\cref{subsec:ablation_analysis}).

\subsection{Experimental Settings}
\label{subsec:settings}

\paragraph{Implementation Details}
We conduct experiments with Stable Diffusion v2.1-base \citep{rombach2022high} on a 20-concept test split from StickerBench in the single-image sticker personalization setting, where each concept is represented by one reference image. We evaluate SEAL on seven TTF personalization baselines: Textual Inversion \citep{gal2022textual}, DreamBooth \citep{ruiz2023dreambooth}, AttnDreamBooth \citep{pang2024attndreambooth}, ClassDiffusion \citep{huang2024classdiffusion}, Custom Diffusion \citep{kumari2023multi}, CoRe \citep{wu2025core}, and UnZipLoRA \citep{liu2025unziplora}. For detailed direct evaluations and ablation studies, we focus on Custom Diffusion, CoRe, and UnZipLoRA. For each baseline, we compare the original method and its module-integrated variant using identical reference images and prompts. Unless otherwise specified, the main comparisons and ablation studies use the 20-concept split. The multi-seed evaluation uses an expanded 200-concept split. The direct per-attribute adherence and background-leakage evaluations also use this expanded split, with 18 structured tag prompts per concept.

All experiments are implemented with the Diffusers \citep{diffusers} library using mixed precision (FP16). Embedding adaptation is run for 250 steps with AdamW using a constant learning rate of $1.5 \times 10^{-4}$ and a batch size of 1. We set the number of auxiliary embeddings in the Split-merge Token Strategy to $K=5$ and merge them at $T_m=100$ adaptation steps. We use $\lambda_{bind}=10^{-3}$, $\lambda_{supp}=1.0$, $\lambda_{spatial}=2\times10^{-3}$, and $\delta=10^{-8}$. For spatial supervision, we generate object masks using the SAM ViT-H checkpoint \citep{kirillov2023segment}. At inference, we use the DPM-Solver++ multistep scheduler \citep{lu2025dpmsolver} with 50 denoising steps and a classifier-free guidance scale of 7.5. For the main comparisons and ablations, we use the same random seed for paired runs.

\paragraph{Evaluation Metrics}
We evaluate identity preservation and prompt alignment under attribute-level prompt edits. Identity preservation is quantified via CLIP-I \citep{radford2021learning} and DINOv2 \citep{oquab2023dinov2} similarities between generated images and the reference image. Prompt alignment is measured by CLIP-T. StickerBench provides structured tags that factorize prompts into multiple attributes (\textit{e.g.}, Appearance, Emotion, Action, Camera Composition, Style, and Background), allowing us to vary contextual attributes while keeping the concept token fixed. This controlled setup enables a more fine-grained assessment of identity preservation and contextual controllability under systematic context changes rather than free-form sentence variations.

\paragraph{Robustness across Random Seeds}
\Cref{tab:multiseed} reports results for SEAL-integrated methods on the 200 single-image concepts using three random seeds.

\begin{table}[t]
    \centering
    \caption{Multi-seed results for SEAL-integrated methods on the 200-concept single-image test split. Values are reported as mean $\pm$ standard deviation over three random seeds.}
    \label{tab:multiseed}
    \resizebox{0.85\textwidth}{!}{%
        \begin{tabular}{l ccc}
            \toprule
            \textbf{Method}
                & \textbf{CLIP-T} $\uparrow$
                & \textbf{CLIP-I} $\uparrow$
                & \textbf{DINOv2} $\uparrow$ \\
            \midrule
            Custom Diffusion + SEAL
                & $0.3195 \pm 0.0021$
                & $0.8592 \pm 0.0078$
                & $0.7071 \pm 0.0118$ \\
            CoRe + SEAL
                & $0.2892 \pm 0.0018$
                & $0.8375 \pm 0.0099$
                & $0.5794 \pm 0.0136$ \\
            UnZipLoRA + SEAL
                & $0.3439 \pm 0.0024$
                & $0.8593 \pm 0.0084$
                & $0.7124 \pm 0.0178$ \\
            \bottomrule
        \end{tabular}
    }
\end{table}

Scores vary little across seeds, indicating limited sensitivity to random initialization. This experiment measures seed stability and does not test statistical differences from the baselines.

\subsection{Comparison with State-of-the-Art Methods}
\label{subsec:comparisons}

\paragraph{Quantitative Analysis}
\Cref{tab:integration_seal} reports results on StickerBench for single-image sticker personalization across metrics for identity preservation and prompt alignment. Our objective is to evaluate the effect of integrating the proposed adaptation module into existing personalization methods rather than proposing a standalone pipeline. We therefore compare baselines and their module-integrated variants.

\begin{table}[t]
    \centering
    \caption{Integration results on StickerBench (one-shot). We report seven baselines and the effect of integrating SEAL into each TTF method. Positive changes over the corresponding baseline are shown in blue parentheses.}
    \label{tab:integration_seal}
    \resizebox{0.85\textwidth}{!}{%
        \begin{tabular}{l cc cc cc}
            \toprule
            \textbf{Method}
                & \multicolumn{2}{c}{\textbf{CLIP-T} $\uparrow$}
                & \multicolumn{2}{c}{\textbf{CLIP-I} $\uparrow$}
                & \multicolumn{2}{c}{\textbf{DINOv2} $\uparrow$} \\
            \cmidrule(lr){2-3} \cmidrule(lr){4-5} \cmidrule(lr){6-7}
                & \textbf{Base} & \textbf{+SEAL}
                & \textbf{Base} & \textbf{+SEAL}
                & \textbf{Base} & \textbf{+SEAL} \\
            \midrule
            Textual Inversion \citep{gal2022textual}
                & 0.274 & 0.293 \textcolor{blue}{(+0.019)}
                & 0.695 & 0.667 (-0.028)
                & 0.399 & 0.454 \textcolor{blue}{(+0.055)} \\
            DreamBooth \citep{ruiz2023dreambooth}
                & 0.298 & 0.309 \textcolor{blue}{(+0.011)}
                & 0.598 & 0.634 \textcolor{blue}{(+0.036)}
                & 0.317 & 0.432 \textcolor{blue}{(+0.115)} \\
            AttnDreamBooth \citep{pang2024attndreambooth}
                & 0.274 & 0.297 \textcolor{blue}{(+0.023)}
                & 0.796 & 0.776 (-0.020)
                & 0.512 & 0.586 \textcolor{blue}{(+0.074)} \\
            ClassDiffusion \citep{huang2024classdiffusion}
                & 0.283 & 0.298 \textcolor{blue}{(+0.015)}
                & 0.769 & 0.795 \textcolor{blue}{(+0.026)}
                & 0.652 & 0.686 \textcolor{blue}{(+0.034)} \\
            \midrule
            Custom Diffusion \citep{kumari2023multi}
                & 0.309 & 0.339 \textcolor{blue}{(+0.030)}
                & 0.835 & 0.865 \textcolor{blue}{(+0.030)}
                & 0.673 & 0.710 \textcolor{blue}{(+0.037)} \\
            CoRe \citep{wu2025core}
                & 0.288 & 0.306 \textcolor{blue}{(+0.018)}
                & 0.915 & 0.849 (-0.066)
                & 0.483 & 0.591 \textcolor{blue}{(+0.108)} \\
            UnZipLoRA \citep{liu2025unziplora}
                & 0.340 & 0.342 \textcolor{blue}{(+0.002)}
                & 0.838 & 0.851 \textcolor{blue}{(+0.013)}
                & 0.667 & 0.705 \textcolor{blue}{(+0.038)} \\
            \bottomrule
        \end{tabular}%
    }
\end{table}

Across the seven evaluated baselines, the module-integrated variants shift the trade-off between identity preservation and contextual controllability. SEAL improves CLIP-T and DINOv2 for all seven methods, while CLIP-I improves for four methods and decreases for Textual Inversion, AttnDreamBooth, and CoRe. Because both CLIP-I and DINOv2 are computed over the full image, they can reward the retention of reference-specific background and composition. We therefore interpret these metrics jointly with the per-attribute adherence and background-only evaluations rather than treating either metric as a direct measure of identity preservation.

\paragraph{Direct Evaluation of Prompt Adherence and Background Leakage}We complement the global CLIP and DINOv2 scores with two evaluations targeting attribute-level prompt adherence and reference-background retention. Both evaluations use the 200-concept test split and 18 structured tag prompts per concept as the main comparison. For each baseline and its SEAL-integrated variant, we evaluate outputs generated from identical reference images, prompts, and the same random seed. Scores are first averaged over the prompts for each concept and then across concepts.

For per-attribute adherence, we use Qwen3-VL-8B-Instruct \citep{bai2025qwen3vl} as a fixed vision-language evaluator. The evaluator receives a generated image and the six requested prompt fields: Appearance, Emotion, Action, Camera Composition, Style, and Background. A single structured evaluation returns a binary judgment for each field. A field is counted as correct only when all requested details in that field are visible in the generated image. We report the proportion of correct judgments for each attribute and their macro average.

\begin{table}[t]
    \centering
    \caption{Per-attribute prompt adherence. Each requested attribute in the prompt is converted into a yes/no question and scored with a vision-language model. We report the proportion of correctly reflected attributes for each category. Positive changes over the corresponding baseline are shown in blue parentheses.}
    \label{tab:attribute_adherence}
    \resizebox{\textwidth}{!}{%
        \begin{tabular}{l cccccc c}
            \toprule
            \textbf{Method}
                & \textbf{Appearance}
                & \textbf{Emotion}
                & \textbf{Action}
                & \textbf{Camera Comp.}
                & \textbf{Style}
                & \textbf{Background}
                & \textbf{Avg.} \\
            \midrule
            Custom Diffusion \citep{kumari2023multi}
                & 0.944
                & 0.325
                & 0.653
                & 0.985
                & 0.971
                & 0.775
                & 0.776 \\
            \quad + SEAL
                & 0.989 \textcolor{blue}{(+0.045)}
                & 0.475 \textcolor{blue}{(+0.150)}
                & 0.665 \textcolor{blue}{(+0.012)}
                & 0.990 \textcolor{blue}{(+0.005)}
                & 0.989 \textcolor{blue}{(+0.018)}
                & 0.910 \textcolor{blue}{(+0.135)}
                & 0.836 \textcolor{blue}{(+0.060)} \\
            CoRe \citep{wu2025core}
                & 0.986
                & 0.350
                & 0.662
                & 0.970
                & 0.938
                & 0.764
                & 0.778 \\
            \quad + SEAL
                & 0.997 \textcolor{blue}{(+0.011)}
                & 0.475 \textcolor{blue}{(+0.125)}
                & 0.565 (-0.097)
                & 0.985 \textcolor{blue}{(+0.015)}
                & 0.962 \textcolor{blue}{(+0.024)}
                & 0.786 \textcolor{blue}{(+0.022)}
                & 0.795 \textcolor{blue}{(+0.017)} \\
            UnZipLoRA \citep{liu2025unziplora}
                & 0.992
                & 0.325
                & 0.438
                & 0.975
                & 0.944
                & 0.700
                & 0.729 \\
            \quad + SEAL
                & 0.997 \textcolor{blue}{(+0.005)}
                & 0.500 \textcolor{blue}{(+0.175)}
                & 0.515 \textcolor{blue}{(+0.077)}
                & 0.970 (-0.005)
                & 0.968 \textcolor{blue}{(+0.024)}
                & 0.778 \textcolor{blue}{(+0.078)}
                & 0.788 \textcolor{blue}{(+0.059)} \\
            \bottomrule
        \end{tabular}%
    }
\end{table}

\Cref{tab:attribute_adherence} shows that integrating SEAL improves the macro-average prompt-adherence score for all three personalization methods. The gains are especially consistent for Emotion and Background, while the per-attribute results also reveal method-specific trade-offs, including lower Action adherence for CoRe and a small decrease in Camera Composition adherence for UnZipLoRA.

\begin{table}[t]
    \centering
    \caption{Background leakage. Similarity is computed between the reference and generated images over background regions only, after excluding the subject; lower values indicate less retention of the reference background. Diversity is one minus the mean pairwise DINOv2 similarity among generated backgrounds for the same concept; higher values indicate greater variation across prompts.}
    \label{tab:background_leakage}
    \resizebox{0.8\textwidth}{!}{%
        \begin{tabular}{l cc}
            \toprule
            \textbf{Method}
                & \textbf{Background similarity} $\downarrow$
                & \textbf{Background diversity} $\uparrow$ \\
            \midrule
            Custom Diffusion \citep{kumari2023multi} & 0.2612 & 0.7197 \\
            \quad + SEAL                             & 0.2382 & 0.7511 \\
            CoRe \citep{wu2025core}                  & 0.2584 & 0.7259 \\
            \quad + SEAL                             & 0.2413 & 0.7823 \\
            UnZipLoRA \citep{liu2025unziplora}       & 0.2136 & 0.7867 \\
            \quad + SEAL                             & 0.2017 & 0.8013 \\
            \bottomrule
        \end{tabular}%
    }
\end{table}

For background leakage, we separate the foreground subject from both the reference and generated images. The reference foreground is obtained from the SAM mask used during adaptation, while each generated subject is segmented independently using SAM with a $3 \times 3$ grid of point-prompt candidates. After excluding the foreground regions, we extract DINOv2 features from the background-only images. Background similarity is computed between the reference background and each generated background, where a lower value indicates less retention of reference-specific background content. Background diversity is defined as one minus the mean pairwise DINOv2 similarity among generated backgrounds for the same concept, where a higher value indicates greater variation across prompts.

\Cref{tab:background_leakage} evaluates reference-background retention after excluding the subject region. This directly measures the background component of \textit{visual entanglement}.

\paragraph{Human Evaluation of Personalization Results}
The same 20 participants evaluated personalization results using 30 pairwise questions from StickerBench. Each question presented one reference image, one target prompt, and two anonymized generation results obtained from a baseline and its corresponding SEAL-integrated variant. For Subject-Background Disentanglement, Contextual Controllability, and Overall Personalization Quality, participants selected the preferred result from each pair. Ties were not allowed, and the presentation order of the two results was randomized for each question.

\begin{table}[t]
    \centering
    \caption{Human evaluation of personalization results by 20 participants using 30 pairwise questions from StickerBench. Each question compares one result from a baseline with one result from its SEAL-integrated variant. The SEAL preference rate is the percentage of pairwise judgments in which participants preferred the SEAL-integrated result; higher values indicate stronger preference for SEAL. Each entry is computed from 200 judgments.}
    \label{tab:user_study}
    \resizebox{0.92\textwidth}{!}{%
        \begin{tabular}{l c ccc}
            \toprule
            \textbf{SEAL-integrated variant}
                & \textbf{Questions}
                & \textbf{Disentanglement (\%)} $\uparrow$
                & \textbf{Controllability (\%)} $\uparrow$
                & \textbf{Overall quality (\%)} $\uparrow$ \\
            \midrule
            Custom Diffusion + SEAL
                & 10
                & 76.5\%
                & 81.0\%
                & 80.5\% \\
            CoRe + SEAL
                & 10
                & 77.5\%
                & 76.5\%
                & 77.5\% \\
            UnZipLoRA + SEAL
                & 10
                & 74.5\%
                & 64.5\%
                & 70.0\% \\
            \bottomrule
        \end{tabular}%
    }
\end{table}

The study contained 10 questions each for Custom Diffusion, CoRe, and UnZipLoRA, yielding 600 judgments per criterion across the 20 participants.

For each baseline and criterion, we report the SEAL preference rate, defined as the percentage of pairwise judgments in which participants preferred the SEAL-integrated result over its corresponding baseline. Each table entry is therefore computed from 200 judgments (20 participants $\times$ 10 questions). A preference rate above 50\% indicates that the SEAL-integrated variant was preferred more often than its baseline.

\begin{figure}[t]
    \centering
    \includegraphics[width=\columnwidth,height=0.38\textheight,keepaspectratio]{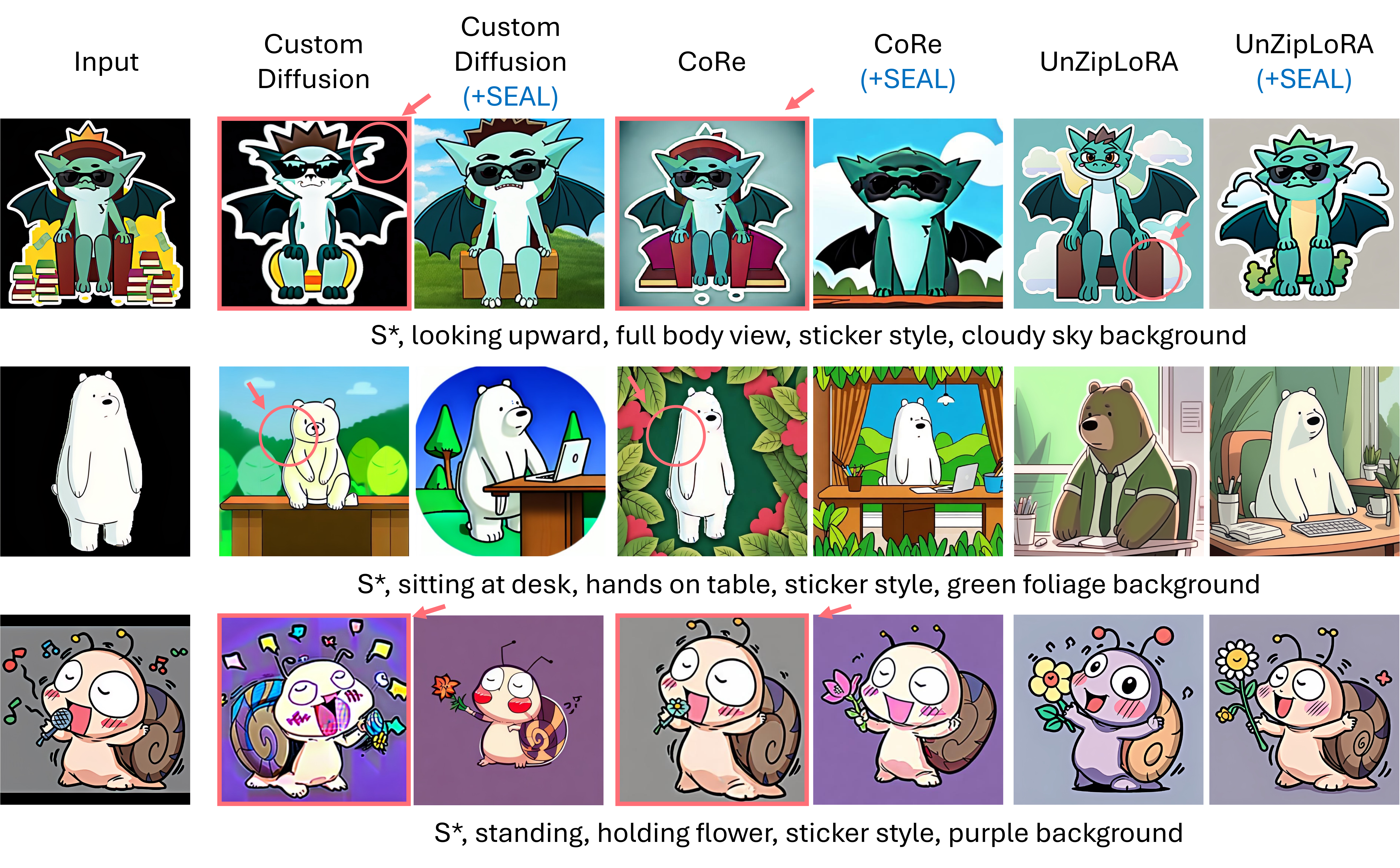}
    \caption{Qualitative comparison of baseline personalization methods and their SEAL-integrated variants in single-image sticker personalization. While baselines often exhibit visual entanglement or structural rigidity, integrating SEAL improves identity disentanglement and contextual controllability under the same prompts.}
    \label{fig:comparison}
\end{figure}

\Cref{tab:user_study} reports the SEAL preference rate in pairwise comparisons with each corresponding baseline. SEAL is preferred in more than 50\% of judgments for every method and criterion, with preference rates ranging from 64.5\% to 81.0\%. Custom Diffusion + SEAL receives the strongest preference for Contextual Controllability (81.0\%), while UnZipLoRA + SEAL shows a smaller but still positive preference (64.5\%) on the same criterion. Preferences for Subject-Background Disentanglement indicate less transfer of reference-background content, while preferences for Contextual Controllability indicate that the subject follows the target prompt without reproducing the pose or composition of the reference image. The Overall Personalization Quality results jointly reflect subject fidelity, target-prompt alignment, and visual coherence.

\begin{figure}[t]
    \centering
    \includegraphics[width=\columnwidth,height=0.38\textheight,keepaspectratio]{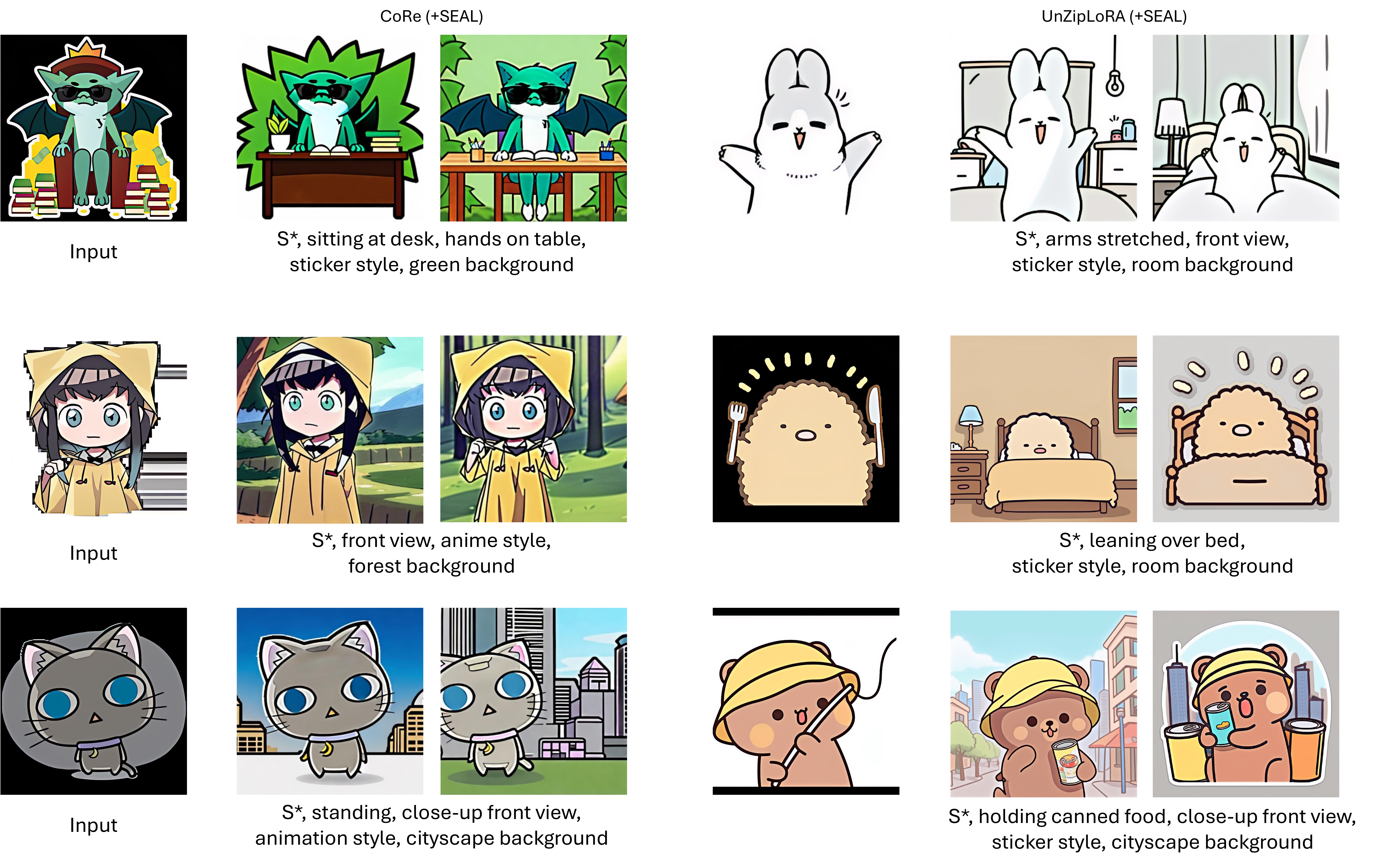}
    \caption{Qualitative results of SEAL integrated into representative TTF personalization methods (CoRe \citep{wu2025core} and UnZipLoRA \citep{liu2025unziplora}) in the one-shot setting. For each concept, the leftmost image is the single reference input, and the remaining images are generated samples conditioned on the prompt shown below each row. Across diverse attribute edits, the module-integrated variants preserve concept identity while maintaining contextual controllability, indicating reduced \textit{visual entanglement} and alleviated \textit{structural rigidity}.}
    \label{fig:main_results}
\end{figure}

When averaged across the three evaluation criteria, participants preferred the SEAL-integrated variants in 79.3\%, 77.2\%, and 69.7\% of the comparisons for Custom Diffusion, CoRe, and UnZipLoRA, respectively.

\begin{figure}[t]
    \centering
    \includegraphics[width=\columnwidth,height=0.38\textheight,keepaspectratio]{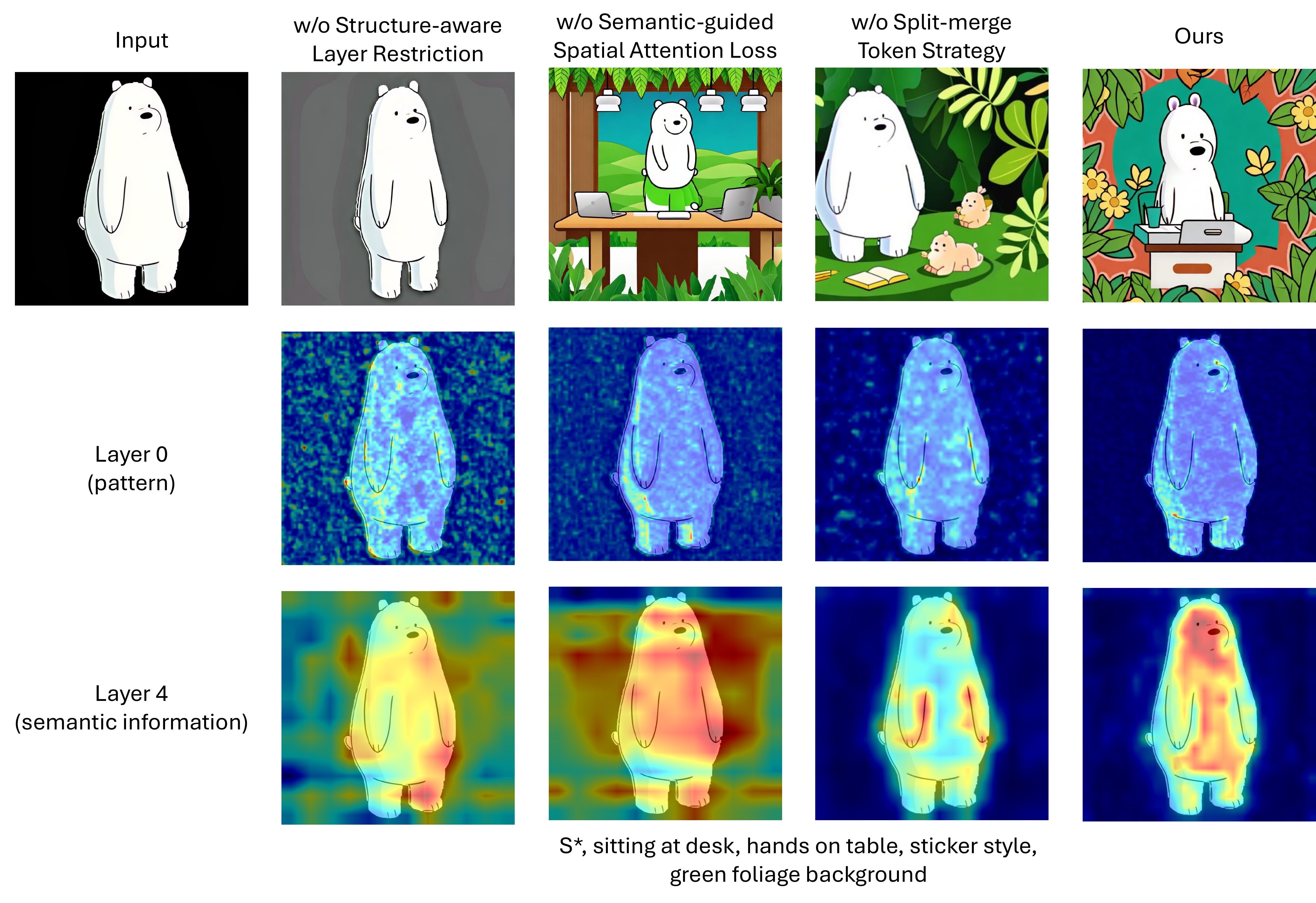}
    \caption{Visual ablation study of SEAL on StickerBench for single-image sticker personalization using CoRe \citep{wu2025core} as the underlying personalization baseline. We show generated samples (top) and cross-attention maps of the concept token (bottom). Removing Structure-aware Layer Restriction increases \textit{structural rigidity}, while removing the Semantic-guided Spatial Attention Loss increases \textit{visual entanglement}. Removing the Split-merge Token Strategy reduces optimization stability under one-shot supervision and degrades identity preservation. The full module improves identity disentanglement and contextual controllability.}
    \label{fig:ablation}
\end{figure}

It is also useful to interpret the results through method-specific characteristics. CoRe \citep{wu2025core} promotes semantic consistency through embedding regularization, which can yield strong identity fidelity but may remain sensitive to background leakage without explicit spatial constraints on the concept token. In contrast, methods such as Custom Diffusion \citep{kumari2023multi} and UnZipLoRA \citep{liu2025unziplora} provide strong adaptation capacity but can still exhibit layout memorization when the adaptation signal is dominated by structurally biased layers. Integrating SEAL addresses these failure modes in a unified manner by constraining where the concept token attends and by emphasizing semantically informative layers during adaptation.

\paragraph{Qualitative Analysis}
\Cref{fig:comparison} presents qualitative comparisons under identical prompts. Standard personalization baselines often exhibit \textit{visual entanglement}, where background elements from the reference image are reflected in the learned concept, or \textit{structural rigidity}, where reference-specific geometry is repeated across prompts. These failures are pronounced in prompts such as ``cloudy sky'' and ``sitting at desk'', where the model must change the environment or composition while preserving identity cues.

\begin{table}[t]
    \centering
    \caption{Component and pairwise ablation of the three SEAL components on StickerBench, using CoRe \citep{wu2025core} as the underlying personalization baseline. SL denotes the Semantic-guided Spatial Attention Loss, ST the Split-merge Token Strategy, and LR the Structure-aware Layer Restriction. Best values are \textbf{bolded}, and second-best values are \underline{underlined}.}
    \label{tab:ablation_detailed}
    \resizebox{0.45\textwidth}{!}{%
        \begin{tabular}{ccc ccc}
            \toprule
            \textbf{SL} & \textbf{ST} & \textbf{LR}
                & \textbf{CLIP-T} $\uparrow$
                & \textbf{CLIP-I} $\uparrow$
                & \textbf{DINOv2} $\uparrow$ \\
            \midrule
            \xmark & \xmark & \xmark & 0.288 & \underline{0.915} & 0.483 \\
            \cmark & \xmark & \xmark & 0.285 & 0.906 & 0.491 \\
            \xmark & \cmark & \xmark & \textbf{0.313} & 0.832 & 0.471 \\
            \xmark & \xmark & \cmark & 0.285 & 0.891 & \underline{0.587} \\
            \midrule
            \cmark & \cmark & \xmark & 0.280 & \textbf{0.916} & 0.428 \\
            \cmark & \xmark & \cmark & \textbf{0.313} & 0.819 & 0.480 \\
            \xmark & \cmark & \cmark & 0.295 & 0.819 & 0.473 \\
            \midrule
            \cmark & \cmark & \cmark & \underline{0.306} & 0.849 & \textbf{0.591} \\
            \bottomrule
        \end{tabular}%
    }
\end{table}

After integrating SEAL, concept boundaries become cleaner and the concept is synthesized with higher contextual flexibility under the same prompts. This behavior is consistent with the Semantic-guided Spatial Attention Loss, which restricts concept-token attention to the object region to mitigate \textit{visual entanglement}, and with Structure-aware Layer Restriction and the Split-merge Token Strategy, which together reduce \textit{structural rigidity} by avoiding over-commitment to reference-specific structure while stabilizing embedding adaptation. \Cref{fig:main_results} provides additional examples across diverse characters and prompt variations, showing that module-integrated personalization preserves identity cues while responding to changes in action, composition, and background in single-image sticker personalization.
\begin{figure}[t]
    \centering
    \includegraphics[width=\columnwidth,height=0.38\textheight,keepaspectratio]{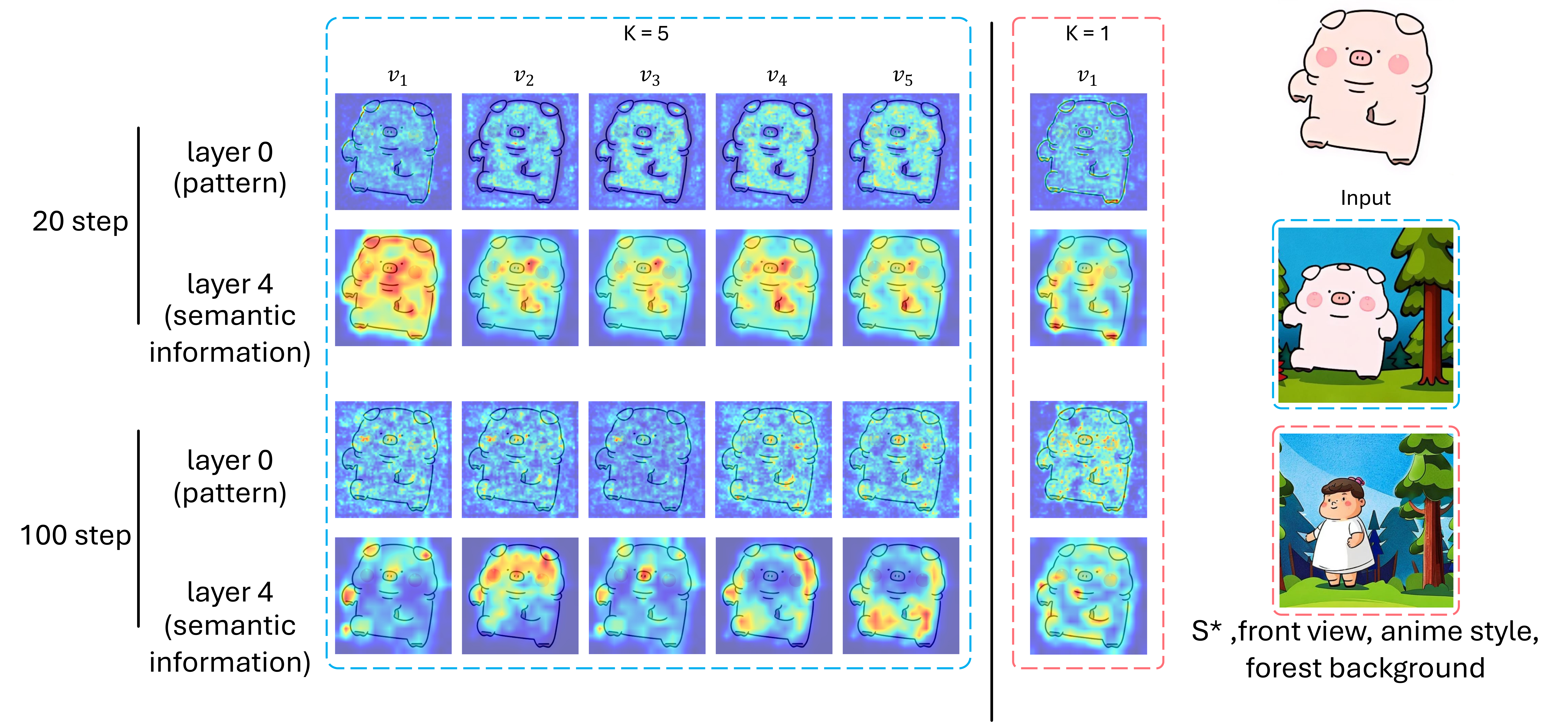}
    \caption{Visual analysis of \textit{structural rigidity} with respect to Structure-aware Layer Restriction during embedding adaptation. Applying spatial supervision to shallow cross-attention layers (\textit{e.g.}, Layer 0), which primarily capture low-level structural patterns, causes the concept embedding to overfit to reference edges, resulting in a fixed layout. In contrast, applying the spatial constraint to deeper, semantically oriented cross-attention layers (\textit{e.g.}, Layer 4) reduces layout memorization and restores contextual controllability under attribute-level prompt edits.}
    \label{fig:k_ablation}
\end{figure}

\subsection{Ablation Studies and Analysis}
\label{subsec:ablation_analysis}

\begin{figure}[t]
    \centering
    \includegraphics[width=\columnwidth,height=0.38\textheight,keepaspectratio]{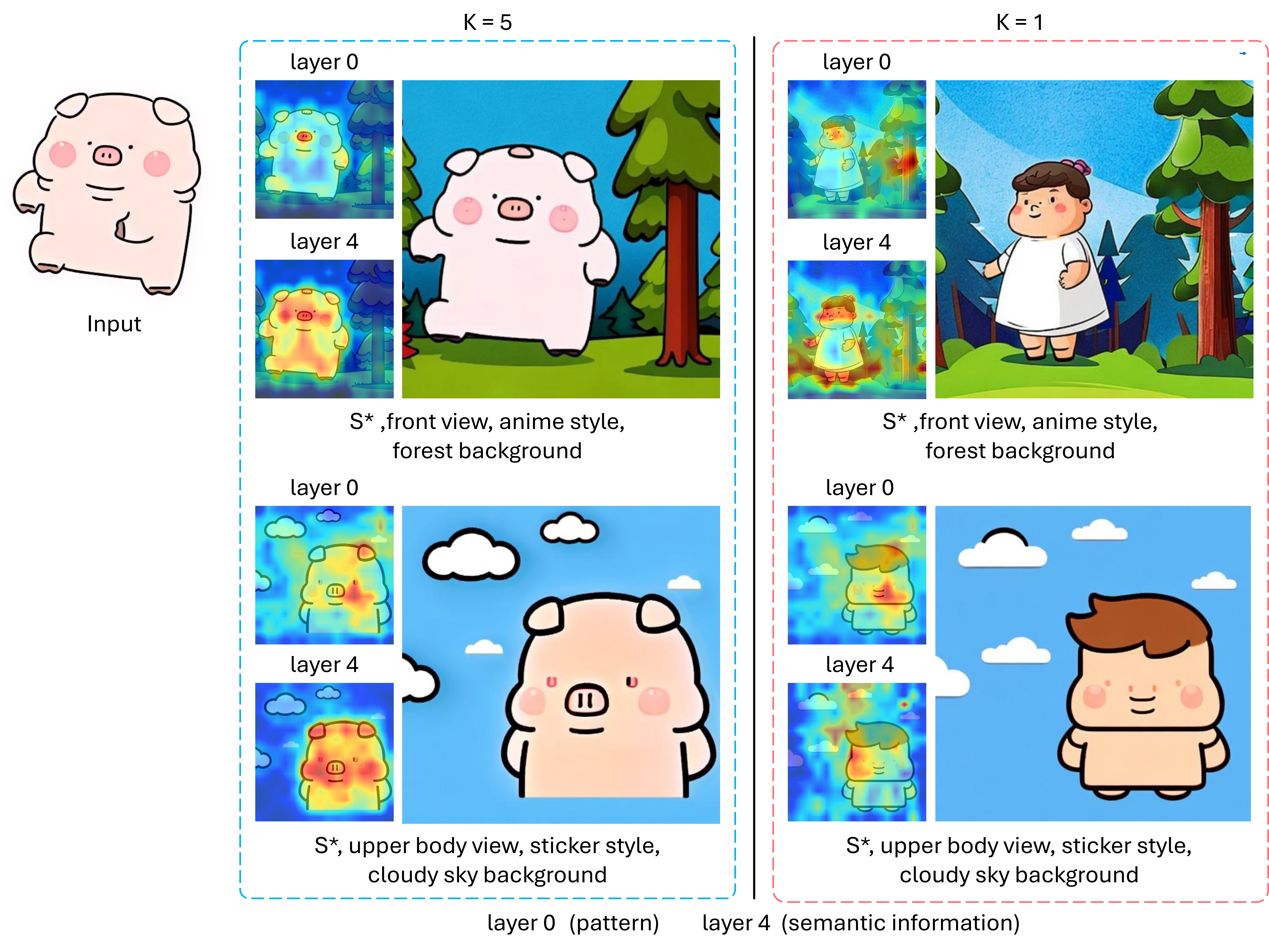} 
    \caption{Inference-time visualization of cross-attention maps across different $K$ values. The visualization uses embeddings learned with the Split-merge Token Strategy. Compared with $K=1$, our selected setting ($K=5$) captures a broader range of concept attributes, yields more coherent attention behavior across layers (\textit{e.g.}, Layers 0 and 4), and better preserves identity under varied prompts.}
    \label{fig:inference_attention}
\end{figure}

\paragraph{Ablation Study on Model Components}
We verify the effectiveness of the proposed adaptation module through visual and quantitative ablations. As shown in \cref{fig:ablation}, removing Structure-aware Layer Restriction causes the adaptation signal to be dominated by image-near cross-attention layers, where attention is strongly influenced by low-level patterns and reference-specific structure. As a result, the adapted concept becomes tied to the reference layout and exhibits severe \textit{structural rigidity}, producing outputs that follow the reference composition even when prompts request changes in action, background, or composition. In contrast, applying the spatial constraint to semantically informative layers prevents the constraint from reinforcing these input-near structural biases, allowing the embedding to capture identity cues while maintaining contextual flexibility.

\Cref{fig:k_ablation} further visualizes this effect across layers: shallow-layer adaptation follows reference-specific edges and layout, whereas restricting optimization to semantically informative layers produces more coherent concept-token attention and better preserves contextual controllability under prompt edits.

\begin{table}[t]
    \centering
    \caption{Ablation on the cross-attention layers receiving spatial supervision. Layer groups are defined by relative depth in the ordered sequence of cross-attention layers. Best results are bolded, and second-best results are underlined.}
    \label{tab:layer_ablation}
    \resizebox{0.85\textwidth}{!}{%
        \begin{tabular}{l l ccc}
            \toprule
            \textbf{Supervised layers}
                & \textbf{Depth range}
                & \textbf{CLIP-T} $\uparrow$
                & \textbf{CLIP-I} $\uparrow$
                & \textbf{DINOv2} $\uparrow$ \\
            \midrule
            Early
                & $[0,\ |L|/2)$
                & \underline{0.2920}
                & \underline{0.8822}
                & 0.5875 \\
            Middle (selected)
                & $[|L|/4,\ 3|L|/4)$
                & \textbf{0.3063}
                & 0.8491
                & \textbf{0.5915} \\
            Late
                & $[|L|/2,\ |L|)$
                & 0.2879
                & 0.8725
                & \underline{0.5914} \\
            All
                & $[0,\ |L|)$
                & 0.2857
                & \textbf{0.8917}
                & 0.5873 \\
            \bottomrule
        \end{tabular}%
    }
\end{table}

We also observe that removing the Semantic-guided Spatial Attention Loss leads to attention leakage into background regions. In this case, concept-token cross-attention maps spread beyond the object boundary and co-activate with background textures that co-occur in the reference image, resulting in \textit{visual entanglement}. \Cref{tab:ablation_detailed} expands the analysis to the baseline, single-component, pairwise, and full configurations. The single-component rows isolate the contribution of each component, while the pairwise rows examine their combined effects. The pairwise configurations produce different metric trade-offs: combining SL and ST yields the highest CLIP-I, while combining SL and LR ties for the highest CLIP-T. The full module yields the highest DINOv2. These results show that the three components affect different aspects of personalization rather than providing uniform additive gains across all metrics.

\paragraph{Layer Selection}
\Cref{tab:layer_ablation} compares early, middle, late, and all-layer spatial supervision using CoRe as the base method.

\begin{table}[t]
    \centering
    \caption{Robustness of SEAL to controlled perturbations of the SAM mask. All conditions use the same 20 concepts, 18 tag prompts per concept, random seed, and remaining training and inference settings. Erosion and dilation use a morphological radius equal to 10\% of the shorter image dimension. Shift translates the mask by 10\% of the image size along both spatial axes, and partial occlusion removes 25\% of the foreground pixels from the right side. Scores are first averaged over prompts for each concept and then across concepts.}
    \label{tab:mask_quality_robustness}
    \resizebox{0.82\textwidth}{!}{%
        \begin{tabular}{l l ccc}
            \toprule
            \textbf{Mask condition}
                & \textbf{Perturbation}
                & \textbf{CLIP-T} $\uparrow$
                & \textbf{CLIP-I} $\uparrow$
                & \textbf{DINOv2} $\uparrow$ \\
            \midrule
            Original SAM      & None              & 0.3063 & 0.8491 & 0.5915 \\
            Erosion           & 10\% strength    & 0.2995 & 0.8434 & 0.5882 \\
            Dilation          & 10\% strength    & 0.3003 & 0.8404 & 0.5811 \\
            Shift             & 10\% per axis    & 0.2903 & 0.8398 & 0.5903 \\
            Partial occlusion & 25\% foreground & 0.2983 & 0.8441 & 0.5893 \\
            \bottomrule
        \end{tabular}%
    }
\end{table}

The supervised depth changes the balance between prompt alignment and image similarity. The middle range achieves the highest CLIP-T and DINOv2 scores, whereas all-layer supervision yields the highest CLIP-I. We select the middle range because it provides the strongest prompt alignment and DINOv2 similarity while avoiding spatial supervision over all cross-attention layers.

\paragraph{Robustness to Controlled Mask Perturbations}
We evaluate the sensitivity of SEAL to controlled perturbations of the SAM mask while keeping all other training and inference settings fixed.

\Cref{tab:mask_quality_robustness} reports the sensitivity of SEAL to controlled perturbations of the SAM mask. Erosion, dilation, shifting, and partial occlusion reduce the three metrics to varying degrees. Dilation produces the largest reduction in DINOv2, whereas shifting produces the largest reductions in CLIP-T and CLIP-I. These results characterize sensitivity to the tested perturbations and do not imply robustness to arbitrary segmentation errors.

\begin{table}[t]
    \centering
    \caption{Ablation study on the split token count $K$ for the Split-merge Token Strategy on StickerBench for single-image sticker personalization, using CoRe \citep{wu2025core} as the underlying personalization baseline. The best results are shown in \textbf{bold}, and the second-best results are \underline{underlined}.}
    \label{tab:ablation_k_tokens}
    \resizebox{0.60\columnwidth}{!}{%
        \begin{tabular}{l c c c}
            \toprule
            \textbf{Split token count $K$} & \textbf{CLIP-T} $\uparrow$ & \textbf{CLIP-I} $\uparrow$ & \textbf{DINOv2} $\uparrow$ \\
            \midrule
            $K = 1$ & \textbf{0.313} & 0.819 & 0.480 \\
            $K = 3$ & \underline{0.308} & 0.833 & \underline{0.483} \\
            \textbf{$K = 5$ (selected)} & 0.306 & \underline{0.849} & \textbf{0.591} \\
            $K = 7$ & 0.290 & \textbf{0.884} & 0.479 \\
            \bottomrule
        \end{tabular}%
    }
\end{table}

\begin{figure}[t]
    \centering
    \includegraphics[width=\columnwidth,height=0.42\textheight,keepaspectratio]{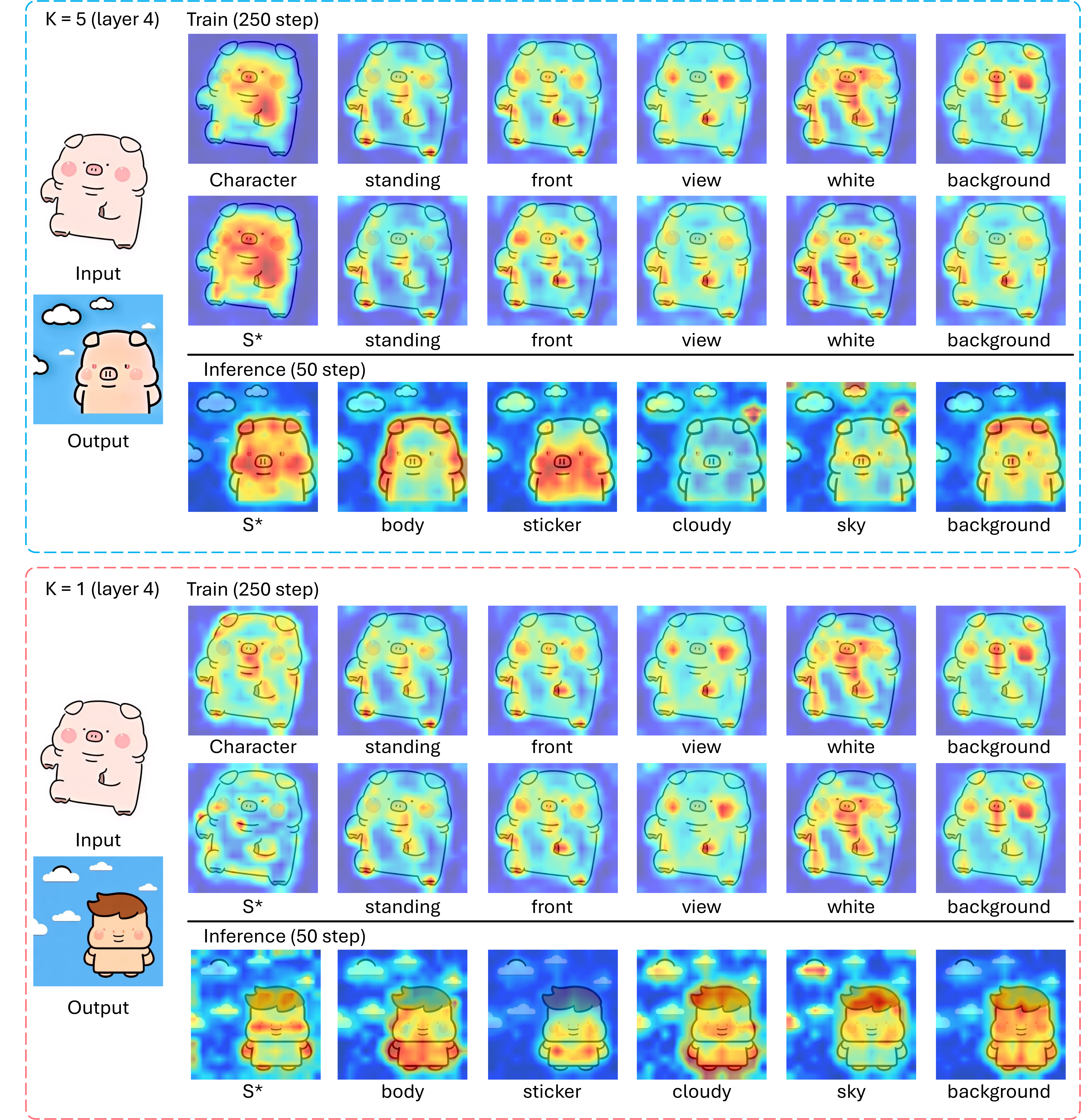}
    \caption{Detailed analysis of cross-attention maps at the end of embedding adaptation (250 steps) and at inference. We compare optimization behavior between $K=5$ (top, blue box) and $K=1$ (bottom, red box) using attention maps from Layer 4. For $K=5$, attention patterns for context tokens remain more consistent when the category token (\textit{e.g.}, ``Character'') is replaced by the concept token $S^*$, indicating more stable semantic injection and improved identity preservation at inference compared with $K=1$.}
    \label{fig:ca_context}
\end{figure}

\paragraph{Split-merge Token Strategy ($K$)}

\begin{figure}[t]
    \centering
    \includegraphics[width=\columnwidth,height=0.38\textheight,keepaspectratio]{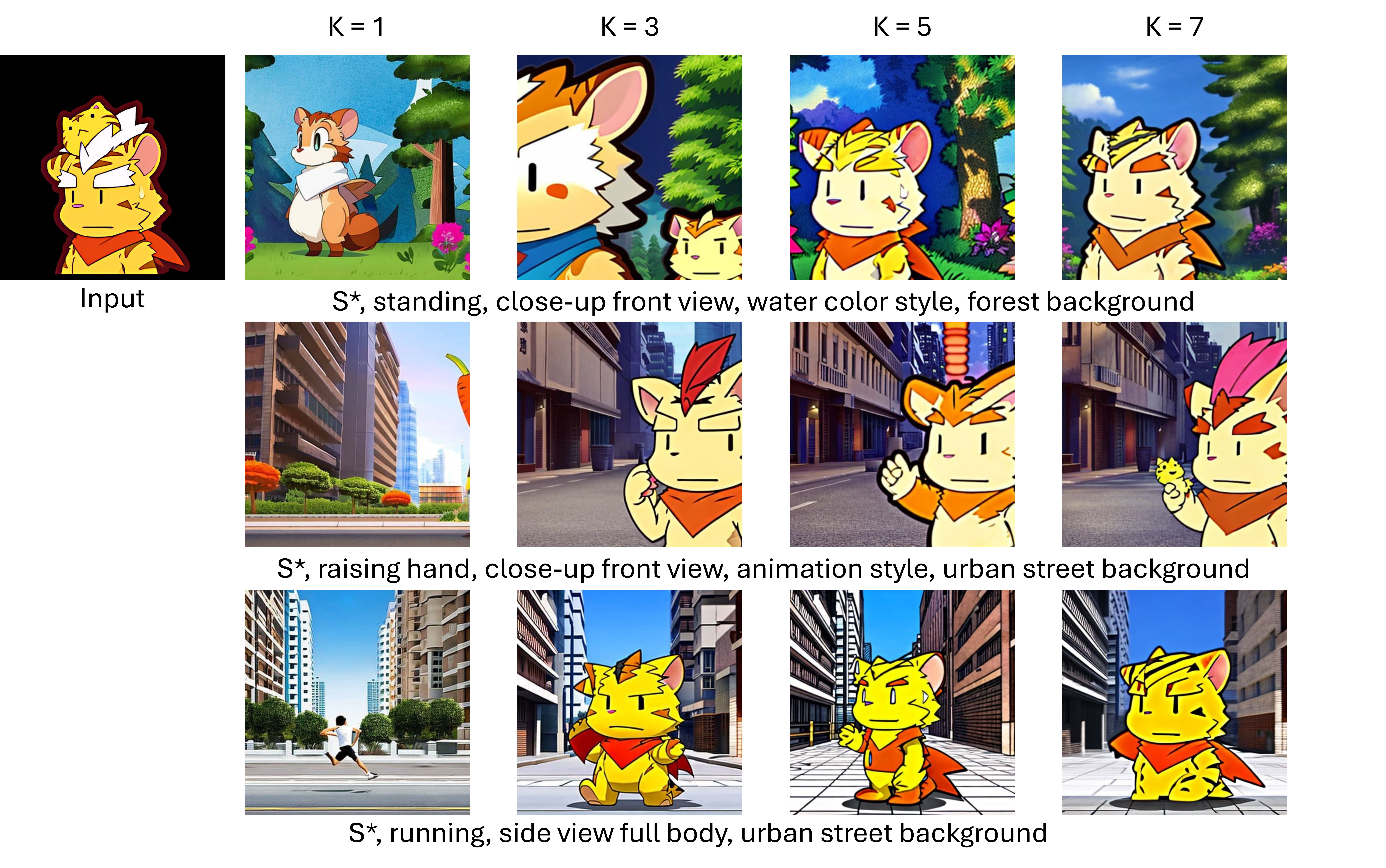}
    \caption{Visual analysis of optimization stability with respect to the number of split tokens $K$. When $K$ is insufficient (\textit{e.g.}, $K=1$), embedding adaptation becomes unstable, leading to degraded identity preservation. In contrast, our Split-merge Token Strategy ($K=5$) improves stability and encourages attribute-diverse concept learning by distributing representation learning across auxiliary embeddings, which helps preserve identity cues while maintaining contextual controllability.}
    \label{fig:k_ablation_2}
\end{figure}

We analyze the influence of the split token count $K$ using the quantitative results in \Cref{tab:ablation_k_tokens}, the inference-time attention maps in \cref{fig:inference_attention}, and the optimization diagnostics in \cref{fig:ca_context} and \cref{fig:k_ablation_2}. With $K=1$, the concept embedding follows a single optimization trajectory and yields lower CLIP-I and DINOv2 scores than $K=5$. This behavior is also reflected in less stable attention patterns and degraded identity preservation under attribute-level prompt edits, even though $K=1$ yields a higher CLIP-T score.

With $K=5$, the auxiliary embeddings receive position-dependent optimization signals before being merged by averaging at step $T_m$, encouraging attribute-diverse concept learning. The resulting embedding produces more consistent concept-token attention behavior across layers and prompts (\cref{fig:inference_attention}), which helps preserve identity cues while maintaining contextual controllability. The diagnostic in \cref{fig:ca_context} shows that context-token attention remains more consistent when a generic category token is replaced by the learned concept token $S^*$. \Cref{tab:ablation_k_tokens} and \cref{fig:k_ablation_2} further show that $K=5$ improves CLIP-I and DINOv2 over $K=1$, although $K=1$ yields a higher CLIP-T score. Increasing $K$ to $7$ yields a different trade-off: CLIP-I increases, but CLIP-T and DINOv2 decrease, suggesting closer matching to the single reference at the expense of prompt adherence and structure-related consistency. DINOv2 does not vary monotonically with $K$, and the three metrics do not peak at the same setting. Because these results come from a single random seed, small differences between neighboring $K$ values should not be over-interpreted. We therefore select $K=5$ as a balance between identity preservation and prompt alignment rather than as the optimum for any single metric.

\paragraph{Computational Cost}
\Cref{tab:compute_cost} compares the computational cost of $K=1$ and $K=5$ using CoRe as the base method while keeping both the embedding-adaptation and U-Net training steps fixed.
\begin{table}[t]
    \centering
    \caption{Computational cost of different split-token settings using CoRe as the base method. Peak GPU memory and wall-clock time are reported as mean $\pm$ standard deviation over three measured concepts.}
    \label{tab:compute_cost}
    \resizebox{0.82\textwidth}{!}{%
        \begin{tabular}{l cc cc}
            \toprule
            \textbf{Configuration}
                & \textbf{Embedding steps}
                & \textbf{U-Net steps}
                & \textbf{Peak GPU Mem. (GB)} $\downarrow$
                & \textbf{Wall-clock time (s)} $\downarrow$ \\
            \midrule
            $K=1$
                & 250
                & 1000
                & $25.62 \pm 0.72$
                & $723.9 \pm 30.2$ \\
            $K=5$ (ours)
                & 250
                & 1000
                & $25.61 \pm 0.72$
                & $745.8 \pm 37.8$ \\
            \bottomrule
        \end{tabular}%
    }
\end{table}

Compared with $K=1$, $K=5$ increases the average wall-clock time from 723.9 to 745.8 seconds, corresponding to an overhead of approximately 3.0\%, while showing no measurable increase in peak GPU memory.

\begin{table}[t]
    \centering
    \caption{Ablation study on prompt representations for adaptation and inference on StickerBench for single-image sticker personalization, using CoRe \citep{wu2025core} as the underlying personalization baseline. We compare sentence templates and structured tags. Best results are shown in \textbf{bold}, and second-best results are \underline{underlined}.}
    \label{tab:ablation_text_form}
    \resizebox{0.60\columnwidth}{!}{%
        \begin{tabular}{l c c c}
            \toprule
            \textbf{Adaptation $\rightarrow$ Inference} & \textbf{CLIP-T} $\uparrow$ & \textbf{CLIP-I} $\uparrow$ & \textbf{DINOv2} $\uparrow$ \\
            \midrule
            Sentence $\rightarrow$ Sentence & 0.261 & \textbf{0.854} & \underline{0.423} \\
            Sentence $\rightarrow$ Tag      & 0.258 & 0.813 & 0.389 \\
            Tag $\rightarrow$ Sentence      & \underline{0.273} & 0.529 & 0.369 \\
            \midrule
            \textbf{Tag $\rightarrow$ Tag} & \textbf{0.306} & \underline{0.849} & \textbf{0.591} \\
            \bottomrule
        \end{tabular}%
    }
\end{table}
\paragraph{Tag-based Prompt Formulations}
We investigate the impact of prompt representations during adaptation and inference in \Cref{tab:ablation_text_form}. Sentence $\rightarrow$ Sentence achieves the highest CLIP-I but substantially lower CLIP-T and DINOv2 scores than Tag $\rightarrow$ Tag. Because CLIP-I measures full-image similarity, its higher score may reflect stronger retention of the reference appearance, background, or composition rather than better personalization under attribute edits. In contrast, Tag $\rightarrow$ Tag achieves the highest CLIP-T and DINOv2 while maintaining competitive CLIP-I, providing a better balance between prompt adherence and reference similarity. The cross-format settings perform worse overall, suggesting that consistency between adaptation-time and inference-time prompt formats is important.

\FloatBarrier

\section{Limitations and Future Work}
\label{sec:limitations}

SEAL relies on an object mask for spatial supervision. Our controlled perturbation analysis considers erosion, dilation, spatial shifting, and partial foreground occlusion, but these conditions do not represent every segmentation failure that may occur in practice. The observed sensitivity to background inclusion and spatial misalignment indicates that the quality of spatial supervision remains important. In addition, Structure-aware Layer Restriction uses a fixed central range of cross-attention layers, and its transferability across different U-Net backbones remains to be studied.

Our study is limited to U-Net-based diffusion models with dedicated token-level cross-attention layers. Multimodal DiT models such as Stable Diffusion 3 \citep{esser2024scaling} and FLUX \citep{flux2024} use joint attention and do not provide the same cross-attention structure or encoder-decoder layer hierarchy assumed by SEAL. Applying SEAL to these models would require different attention extraction and layer selection mechanisms. We therefore consider DiT-based personalization outside the scope of this work and leave it for future study.

\section{Conclusion}
\label{sec:conclusion}

In this paper, we have addressed the challenges of single-image sticker personalization, focusing on two dominant overfitting symptoms: \textit{visual entanglement} and \textit{structural rigidity}. We have proposed SEAL, a plug-and-play semantic adaptation module that can be integrated into test-time fine-tuning (TTF) personalization pipelines without modifying the diffusion backbone. SEAL combines three components: (1) the Semantic-guided Spatial Attention Loss, which suppresses background leakage in concept-token attention to mitigate \textit{visual entanglement}; (2) the Split-merge Token Strategy, which stabilizes embedding adaptation and encourages attribute-diverse concept learning to alleviate \textit{structural rigidity}; and (3) the Structure-aware Layer Restriction, which applies spatial constraints to semantically informative cross-attention layers to further reduce \textit{structural rigidity} and restore contextual controllability. To support systematic study in the sticker domain, we have introduced StickerBench, a large-scale sticker image dataset with structured tags that enable attribute-level prompt edits for evaluating identity disentanglement and contextual controllability. Quantitative and qualitative results have shown that integrating SEAL improves the balance between identity preservation and contextual controllability in single-image sticker personalization. These findings highlight the importance of explicit spatial and structural constraints during embedding adaptation for reliable personalization from a single reference image.

\section*{Acknowledgments}

This work was partly supported by Institute of Information \& communications Technology Planning \& Evaluation (IITP) grant funded by the Korea government(MSIT) (No. RS-2025-25422680, Metacognitive AGI Framework and its Applications, 50\%) This research was supported by Culture, Sports and Tourism R\&D Program through the Korea Creative Content Agency grant funded by Ministry of Culture, Sports and Tourism in 2024 (Project Name : Developing Professionals for R\&D in Contents Production Based on Generative Ai and Cloud, Project Number : RS-2024-00352578, Contribution Rate: 50\%).


\bibliography{sec/main} 

\end{document}